\documentclass[10pt,twocolumn,letterpaper]{article}

\usepackage[pagenumbers]{cvpr} 

\definecolor{cvprblue}{rgb}{0.21,0.49,0.74}
\usepackage[pagebackref,breaklinks,colorlinks,allcolors=cvprblue]{hyperref}
\usepackage{fix-cm}
\usepackage{amsmath,amssymb,amsfonts}
\usepackage{textcomp}
\usepackage{dsfont}
\usepackage{bm}
\usepackage{siunitx}
\usepackage[nolist,nohyperlinks]{acronym}
\usepackage{amsmath}
\usepackage{subcaption}
\usepackage{gensymb}
\usepackage{caption}
\usepackage{multirow}
\usepackage{algpseudocode}
\usepackage{algorithm}
\acrodef{POV}[POV]{Point Of View}
\acrodef{ROI}[ROI]{Region Of Interest}
\acrodef{DLT}[DLT]{Direct Linear Transform}
\acrodef{FOV}[FOV]{Field Of View}
\acrodef{PCA}[PCA]{Principal Component Analysis}
\acrodef{COR}[COR]{Coefficient Of Restitution}
\acrodef{PnP}[PnP]{Perspective-n-Point}
\acrodef{ODE}[ODE]{Ordinary Differential Equation}
\acrodef{MAE}[MAE]{Mean Absolute Error}
\acrodef{MRAE}[MRAE]{Mean Relative Absolute Error}
\acrodef{mIoU}[mIoU]{mean Intersection over Union}
\acrodef{GRU}[GRU]{Gated Recurrent Unit}
\acrodef{MLP}[MLP]{Multi-Layer Perceptron}
\acrodef{SNR}[SNR]{Signal Noise Ratio}

\def\paperID{1} 
\def\confName{CVPR}
\def\confYear{2025}

\title{TT3D: Table Tennis 3D Reconstruction}

\author{Thomas Gossard \thanks{This research was funded by Sony AI.} \qquad Andreas Ziegler \qquad Andreas Zell \\
Cognitive Systems - Univeristy of Tuebingen \\
Sand 1 Tuebingen 72076 Germany\\
{\tt\small thomas.gossard@uni-tuebingen.de}
}

\begin{document}
\maketitle
\begin{abstract}
Sports analysis requires processing large amounts of data, which is time-consuming and costly. 
Advancements in neural networks have significantly alleviated this burden, enabling highly accurate ball tracking in sports broadcasts. 
However, relying solely on 2D ball tracking is limiting, as it depends on the camera's viewpoint and falls short of supporting comprehensive game analysis.
To address this limitation, we propose a novel approach for reconstructing precise 3D ball trajectories from online table tennis match recordings. 
Our method leverages the underlying physics of the ball's motion to identify the bounce state that minimizes the reprojection error of the ball's flying trajectory, hence ensuring an accurate and reliable 3D reconstruction. 
A key advantage of our approach is its ability to infer ball spin without relying on human pose estimation or racket tracking, which are often unreliable or unavailable in broadcast footage. 
We developed an automated camera calibration method capable of reliably tracking camera movements.
Additionally, we adapted an existing 3D pose estimation model, which lacks depth motion capture, to accurately track player movements.
Together, these contributions enable the full 3D reconstruction of a table tennis rally.
Project page: \url{https://cogsys-tuebingen.github.io/tt3d/}
\end{abstract}    
\section{Introduction}
\label{sec:intro}
\begin{figure}[h]
    \centering
    \includegraphics[width=0.85\linewidth]{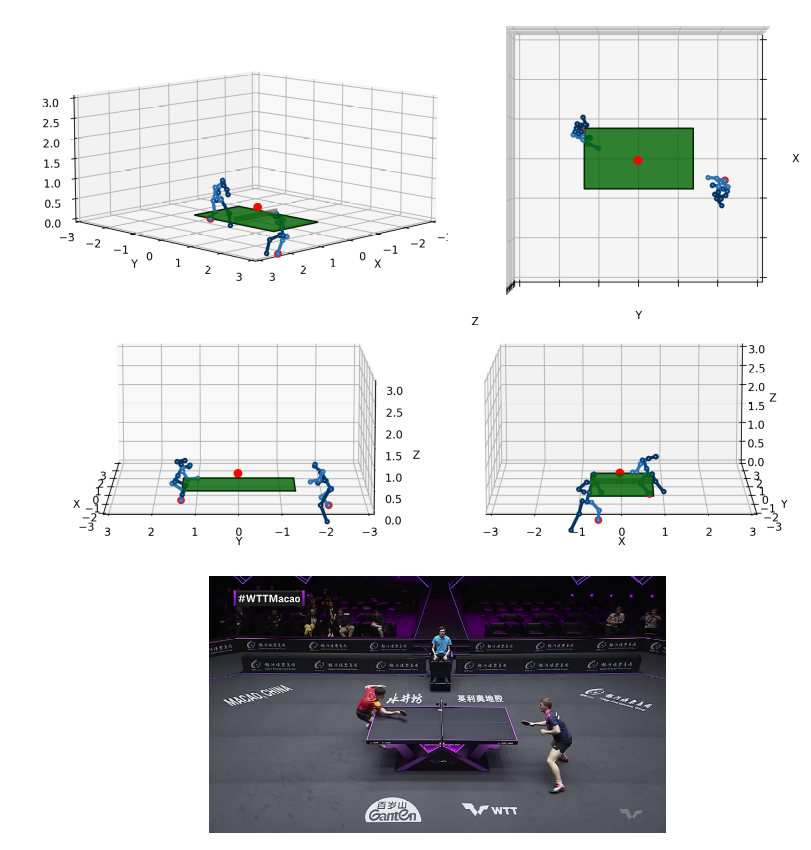}
    \caption{(Top) 3D Reconstruction of table tennis game. We show the reconstruction from different points of view with the players and the ball (red dot). (Bottom) Corresponding frame from real footage.}
    \label{fig:reconstruction}
\end{figure}
%
%
%
%

Accurately capturing the state of a game is fundamental for effective sports analytics. 
Traditionally, this task required extensive manual annotation, making it labor-intensive and time-consuming. 
However, recent advancements in machine learning have significantly automated this process, providing powerful tools to analyze gameplay and extract meaningful insights~\cite{naik2022}. 
Learning-based methods~\cite{zheng2023} allow for the precise extraction of player poses, while advanced object detectors enable effective tracking of the ball and other key objects~\cite{kamble2019}.
By leveraging these automated techniques, raw data can be systematically processed to evaluate player performance, uncover tactical patterns, and refine strategic decision-making, ultimately offering a competitive edge.
This processed data can have multiple uses.
For example, motion-captured recordings of table tennis games were used to find the important shot characteristics necessary to win a rally~\cite{muelling2014c}.
Wu et al. also showed that it is possible to predict the ball bouncing position from the serve stroke motion~\cite{wu2019}.
Using such data can furthermore enable a policy to learn sport skills like tennis strokes ~\cite{zhang2023vid2player3d}.
Finally, precise game state estimation has the potential to enable automated officiating, reducing human error in refereeing and improving fairness in sports~\cite{wong2023}.

Despite these advancements, a key challenge remains: most game recordings are captured using monocular cameras, leading to a loss of depth information.
While 2D ball tracking provides valuable insights, it remains inherently limited by perspective distortions and camera positioning. 
For a more robust and comprehensive analysis, spatial data must be reconstructed in 3D, as it eliminates viewpoint dependencies and allows for a more accurate representation of ball motion and interactions. 
Therefore, the primary objective of this paper is to achieve 3D reconstruction of sports games, with a specific focus on the challenging task of reconstructing ball trajectories for table tennis.

Table tennis presents unique challenges due to its highly dynamic nature. 
For comparison, tennis averages 1.3 hits per second~\cite{zotero-2360}, while badminton ranges from 1.26 to 1.76 hits per second~\cite{laffaye2015, a2017}.
In contrast, table tennis can reach up to 2 hits per second~\cite{a2017}.
Additionally, ball spin plays a crucial role in shaping its trajectory. 
The Magnus effect causes the ball to curve in mid-air, while its spin significantly influences its bounce trajectory, altering its direction upon impact.
These factors make the ball's 3D trajectory particularly complex, as the reconstruction must accurately account for aerodynamic forces and bounce dynamics compared to badminton.
But these dynamics are also what will help infer the 3D motion of the ball from just the 2D observation.
Finally, we have to deal with a lot of ball occlusion due to the players, depending on the point of view. 

In this work, we propose a complete pipeline to reconstruct the 3D state of the table tennis game: 3D player pose and 3D ball trajectory.
For this purpose, we developed an automated camera calibration method capable of tracking both the camera's pose and focal length.
It achieves this by locating the table corners using a segmentation mask of the table surface.
We also propose a novel method to reconstruct 3D ball trajectories from monocular table tennis match recordings.
Our approach leverages the physics of ball motion to reconstruct the ball's 3D trajectory from the table bounce position, achieving robust and accurate predictions even under challenging real-world conditions.
Integrating 3D human pose estimation finally allows us for a complete 3D reconstruction of table tennis matches. 
\section{Related Work}
\label{sec:related_work}
Many 3D reconstruction methods have been developed for different sports such as badminton~\cite{liu2022b}, football~\cite{metzler2013}, tennis~\cite{ertner2024}, basketball~\cite{zandycke2022, chen2009}, table tennis~\cite{shen2016, calandre2021a, tamaki2013} or volleyball~\cite{chen2012}.
Reconstructing the 3D state of a ball sport involves three fundamental components: camera calibration, ball detection and tracking, and 2D-to-3D lifting.
Each step plays a critical role in ensuring accurate 3D trajectory reconstruction.

\subsection{Single View Camera Calibration}

Camera calibration is essential to determine the camera matrix $\bm{K}$ and the camera pose, which is defined by the rotation matrix $\bm{R}$ and the translation vector $\bm{T}$.
These parameters enable the transformation between 3D world coordinates and 2D image coordinates using the pinhole camera model.
Specifically, given a 3D world point $\bm{X_w} = [X, Y, Z]^T$ and its corresponding image point $\bm{x} = [u, v, 1]^T$, the relationship is defined by the projection matrix $\bm{P}$, expressed as
\begin{equation}
\bm{x} = \bm{P} \bm{X_w} = \bm{K} [\bm{R} | \bm{T}] \bm{X_w}. 
\label{eq:camera_model}
\end{equation}
Typically, camera calibration is conducted using a moving calibration pattern, such as a checkerboard, an asymmetric circle grid, or Aprilgrid~\cite{apriltag}.
Broadcast footage usually does not require camera calibration, and information about the camera's placement or focal length is generally not provided in such recordings.
To address this, objects of known geometry in the scene are used for calibration.

In racket sports like badminton and tennis, the intersections of court lines and net poles are commonly used~\cite{xinguoyu2009, liu2022b}.
This process relies on a combination of computer vision techniques, including color filtering, Canny edge detection, Hough line transforms, and sport- or \ac{POV}-specific heuristics.
Currently, no automatic calibration methods exist for table tennis.
This is due to the limited number of field features compared to sports like tennis or badminton, the smaller court size, variations in field color schemes, diverse recording \ac{POV}s, and frequent occlusions.
As a result, many approaches resort to manually annotating features on the images for the calibration process~\cite{ertner2024, calandre2021a, shen2016, wong2023, tamaki2013}.

\subsection{Ball Tracking}

Once the camera is calibrated, the next step is ball detection and tracking.
Traditional computer vision methods, including color filtering, background subtraction, and blob detection, have been widely used for this task~\cite{archana2015, hung2018, tebbe2019, wong2023, tamaki2013}.
More recently, deep learning-based approaches, such as fine-tuned general object detectors like Detectron2~\cite{calandre2021a} and YOLOv4~\cite{kulkarni2022}, have gained popularity in sports applications.
However, in scenarios where a single ball is present, a common approach is to generate a heatmap representing its most probable location using models such as DeepBall~\cite{huang2019}, TracketNet and its variants~\cite{huang2019, sun2020, chen2024, raj2024}, Monotrack~\cite{liu2022b}, WASB~\cite{tarashima2023}, and BlurBall~\cite{blurball}.
Among these, BlurBall not only estimates the ball's position but also provides motion blur information, offering valuable insight into velocity.
This is particularly beneficial in table tennis, where fewer frames are available to capture ball trajectories compared to other racket sports.
In our pipeline, we adopt BlurBall as the primary ball detector.


\subsection{3D Reconstruction}
With the 2D ball positions available, we now need to reconstruct the 3D data from it.
The first step is to segment the whole rally into individual ball trajectories, i.e. between two player strikes.
Previous methods either use heuristics on the change in velocity~\cite{calandre2021a, shen2016} or \ac{GRU} networks to detect frames where the ball bounces or is hit by players~\cite{liu2022b, ertner2024}, using inputs such as field coordinates, ball positions in image space, and player pose.
With the individual trajectory isolated, we can infer the 3D trajectory.

Calandre et al.~\cite{calandre2021a} propose the use of the observed diameter of the ball to infer its 3D position, with depth estimated using the camera's intrinsic parameters. 
The resulting 3D positions are then projected onto a 2D plane fitted to all observed ball locations from the trajectory.
This approach assumes that no sidespin is applied to the ball.
Unfortunately, this method is impractical for broadcast recordings, as the ball's small apparent diameter (lower than 10 pixels) and the sensitivity of depth regression to measurement errors make accurate 3D position estimation unfeasible.
Nonetheless, humans are still able to infer the ball's 3D trajectory to some extent.
This ability stems from our understanding of the physics governing the ball's flight and bounce, which constrains the predicted 3D trajectory only to plausible paths.
The flight path of the ball can be modeled using an \ac{ODE}, where the trajectory is predicted from an initial position and velocity.
By projecting the predicted trajectory onto the image plane using the pinhole camera model, the ball’s initial state can be optimized to minimize the reprojection error. 
Liu et al. employed this approach for badminton~\cite{liu2022b}.
However, the optimization problem is non-convex and prone to local minima, requiring good initialization or additional constraints to guide convergence toward the correct solution.
Liu et al. addressed this issue in badminton by constraining the initial and final 3D positions of the shuttlecock, ensuring they were close to the player's hand as estimated by a pose detection model. 
A similar approach was tested in table tennis~\cite{shen2016} but the additional constraints were set as the ball's first and second bouncing position for the serve.
Indeed, the 3D position of the bouncing ball can be inferred as a ray-plane intersection. 
However, the defined \ac{ODE} completely ignores drag and the Magnus effect and this method was only tested for serves.
An alternative to optimization-based methods is treating the problem as a regression task. 
In tennis, SynthNet~\cite{ertner2024} utilizes a synthetic dataset of ball trajectories to train a \ac{MLP} that predicts the ball’s initial position and velocity based on its 2D trajectory before the bounce and the court’s pose.
\section{Camera Calibration}
\label{sec:camera_calibration}

Camera calibration requires matching 2D image features to their corresponding 3D world points.
In table tennis, the table’s standardized dimensions make it the only viable calibration target.
The world frame is thus defined relative to the table, as illustrated in \cref{fig:table_cal}.

A closed-form solution for estimating camera parameters can be obtained using \ac{DLT} if at least six non-coplanar points are available.
For the table, potential calibration features include its corners (4 points), center line (2 points), and net poles (2 points).
However, these features are often obstructed by players or not detected, making the use of \ac{DLT} impractical.
To simplify the problem, we assume no lens distortion ($\bm{d} = 0$), the optical center aligns with the image center ($c_x = w/2, c_y = h/2$), and the focal lengths are equal in both directions ($f_x = f_y$).
Under these assumptions, the problem falls into the \ac{PnP} domain with unknown focal length (PnPf).
Only 3.5 points are required to estimate the camera's focal length $f$ and extrinsic parameters $\bm{R}$ and $\bm{T}$~\cite{wu2015}. 
As such, we can calibrate the camera with only 4 points (which can be coplanar).
Since it is unlikely for an entire table edge to be obstructed, we chose the table corners—defined as the intersections of the table edges—as our calibration points.

In the following sections, we discuss the various components of the calibration process.
\Cref{ssec:calibration} details the camera calibration approach, assuming we already have access to the necessary keypoints.
Next, we introduce our table tennis table segmentation model in \Cref{ssec:table_segmentation}.
Finally, in \cref{ssec:feature_extraction}, we describe how the table mask is used to extract the features necessary for calibration.

\subsection{Calibration}
\label{ssec:calibration}

Existing PnPf methods are highly complex and depend on advanced algebraic solvers, such as Gröbner solvers~\cite{zheng2014,zheng2016a,wu2015,penatesanchez2013,kanaeva2015}.
Zheng et al.~\cite{zheng2016a} showed that, with a good initialization of the camera pose and focal length, minimizing the reprojection error can achieve accuracy comparable to state-of-the-art PnPf methods.
Given this insight, we opted to frame the PnPf problem as an optimization problem, resulting in a much simpler approach.
This is feasible because the camera is always positioned within a specific distance from the table and must cover the entire field, significantly constraining the range of possible focal lengths.
We describe the optimization procedure in  \Cref{alg:optimize_focal_length}.
We begin by using PnP to estimate the camera pose ($\bm{R}$ and $\bm{T}$) with the currently estimated focal length $f$.
Next, we refine $f$ by minimizing the reprojection error.
These two steps are repeated iteratively until the focal length converges.
As mentioned earlier, proper initialization is crucial.
Manual annotation of features using DLT yielded focal lengths ranging from 1000 to 3000.
Therefore, we initialize the focal length at 1500.
The advantage of our approach is the flexibility. 
It can work with just 4 coplanar points.

\begin{algorithm}[]
\caption{Estimate Focal Length By Minimizing Reprojection Error}
\label{alg:optimize_focal_length}
\begin{algorithmic}[1]
\Require Set of 3D points $\mathbf{X} = \{\bm{X}_i\}$, corresponding 2D image points $\mathbf{x} = \{\bm{x}_i\}$, initial focal length $f_0$, convergence threshold $\epsilon$, maximum iterations $N$
\Ensure Optimized focal length $f^*$

\State Initialize $f \gets f_0$
\State Set iteration counter $k \gets 0$

\Repeat
    \State Solve the PnP problem with $f$ to estimate $\mathbf{R}, \mathbf{T}$
    \State Compute the total reprojection error:
    \[
    E(f) = \sum_i ||\bm{x}_i - \bm{K} [\bm{R}|\bm{T}] \bm{X}_i||
    \]
    \State Update $f$ by minimizing $E(f)$ using a line search
    \State Increment iteration counter $k \gets k + 1$
\Until{$|E(f_{\text{prev}}) - E(f)| < \epsilon$ or $k \geq N$}

\State Set $f^* \gets f$
\State \Return $f^*$
\end{algorithmic}
\end{algorithm}

\begin{figure}
    \centering
    \includegraphics[width=0.9\linewidth]{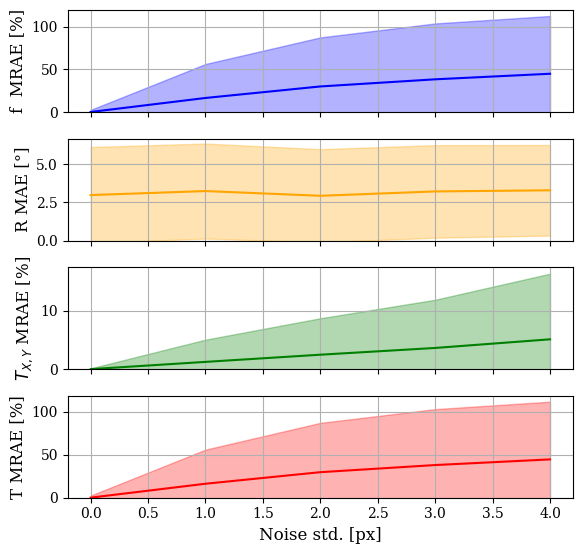}
    \caption{Mean Relative Absolute Error (MRAE) and Mean Absolute Error (MAE) for camera parameters estimated with Algorithm 1 for 1000 samples with noise. The table position was randomly sampled within the camera frame coordinates, ranging from [-2, -2, 5] to [2, 2, 20] (m), and the focal length was randomly sampled between 500 and 3000.}
    \label{fig:sim_cam_calibration}
\end{figure}

To validate our approach, we conducted tests using synthetically generated points for random table positions, orientations, and focal lengths to assess the calibration accuracy of the calibration.
As shown in \Cref{fig:sim_cam_calibration}, the table's rotation is accurately estimated, with an average error of only 3 degrees.
However, the translation vector and focal length show significant errors.
This limitation arises from using only four planar points, which results in a strong correlation between focal length and depth estimation.
This correlation makes it challenging to estimate these parameters independently.
Despite this limitation, it doesn't impact the downstream reconstruction task, as errors in one parameter are often compensated for by corresponding adjustments in the other.
The method demonstrates strong performance in estimating the XY components, achieving a \ac{MRAE} of just 5\%, indicating its particular effectiveness in the plane of the table.



The described method can be applied to individual frames.
However, when processing a video with potential camera movement and zooming, integrating a Kalman Filter can help refine the raw observations, providing smoother and more accurate estimates.

\subsection{Segmentation Model}
\label{ssec:table_segmentation}

To extract the features necessary for the camera calibration, we need to accurately identify the table.
We propose to do so with a segmentation model.
Promptable segmentation models like SAM~\cite{ravi2024sam2segmentimages} can identify table regions, even with players occluding parts of the table.
However, they struggle with net segmentation and often include the table's vertical sides in the mask.
The same goes for depth models like DepthAnything V2~\cite{depth_anything_v2} which do not perfectly allow to identify the table's surface.

To address our specific requirements, we developed a custom table segmentation model.
We began by assembling a dataset of 1,200 images sourced from online videos and manually labeled them.
We then fine-tuned pre-trained models for table tennis table segmentation.
Given the task's relative simplicity, we focused on lightweight models with no more than 2 million parameters.
After extensive testing, we selected a U-Net++ architecture~\cite{zhou2018} with a pre-trained RegNetY encoder~\cite{radosavovic2020} (2M parameters) for its optimal balance between inference speed and accuracy.
The model was trained using the Dice loss~\cite{dice}.
Our final model achieves a \ac{mIoU} of 0.92 on the test set and runs at 70 frames per second, making it suitable for real-time applications.

\subsection{Feature extraction}
\label{ssec:feature_extraction}
\begin{figure}
    \centering
    \begin{subfigure}[b]{0.48\linewidth}
        \includegraphics[width=\textwidth]{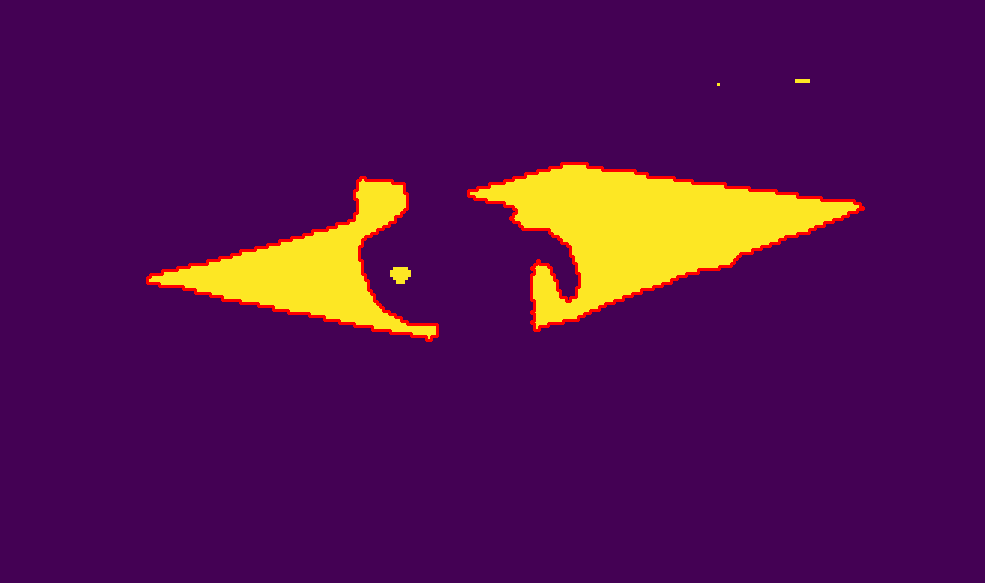}
        \caption{Mask (yellow) and contours (red)}
        \label{fig:table_mask}
    \end{subfigure}
    \hfill
        \begin{subfigure}[b]{0.48\linewidth}
        \includegraphics[width=\textwidth]{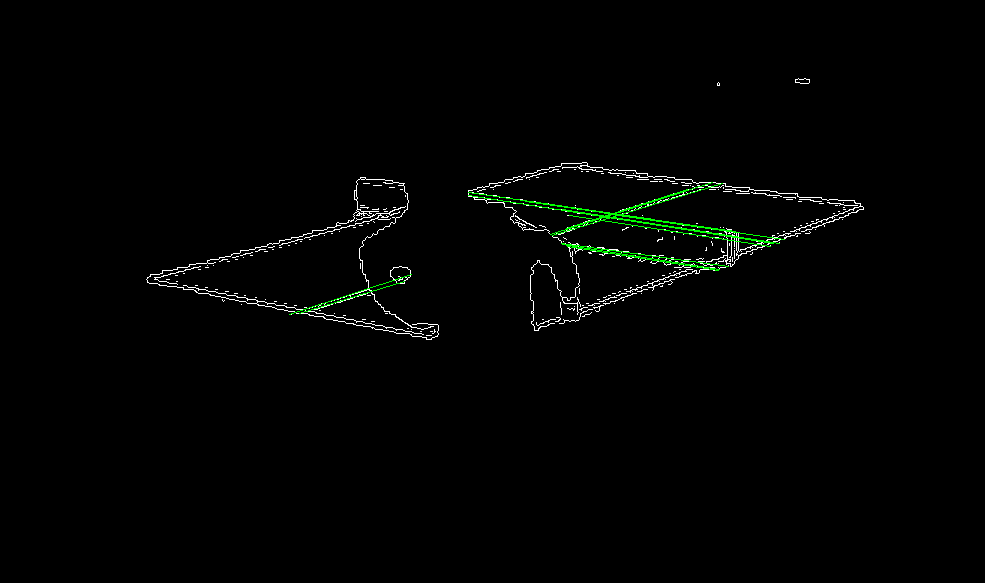}
        \caption{Canny edge within table mask (white) and Hough lines (green)}
        \label{fig:table_midline}
    \end{subfigure}


        \begin{subfigure}[b]{0.48\linewidth}
        \includegraphics[width=\textwidth]{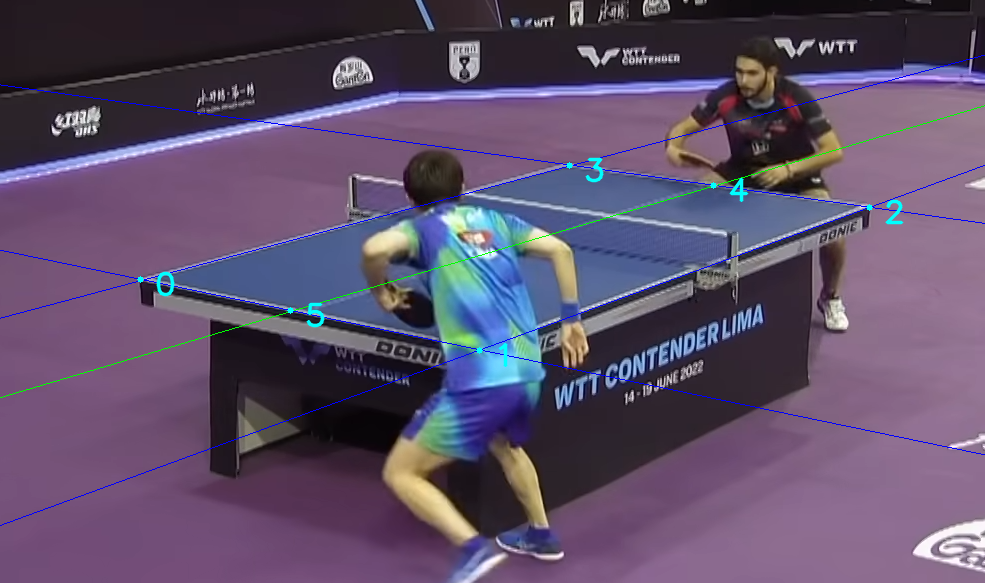}
        \caption{Table edges (blue lines), midline (green), and detected features (numbered points)}
        \label{fig:table_corners}
    \end{subfigure}
    \hfill
    \begin{subfigure}[b]{0.48\linewidth}
        \includegraphics[width=\textwidth]{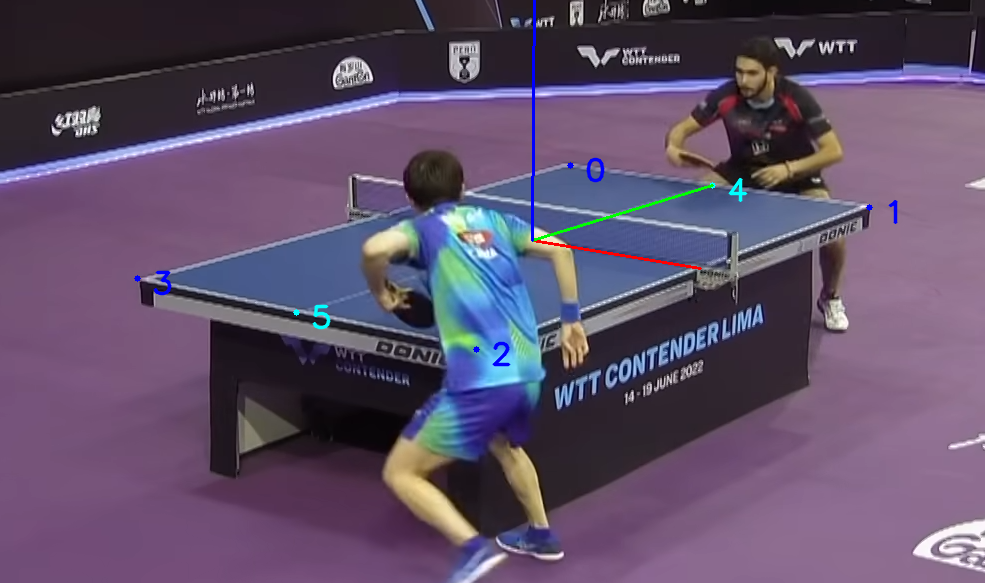}
        \caption{Calibrated camera with reprojected features.Red, Green and blue axes are respectively X, Y and Z.}
        \label{fig:table_cal}
    \end{subfigure}

    \caption{Calibration pipeline. (a) Mask with contour, (b) Midline detection (c) Detected corners and midline, (d) Final calibrated camera.}
    \label{fig:calibration_pipeline}
\end{figure}

We first try to extract the table corners.
We start by extracting the table contours from the segmentation mask using the method proposed by Suzuki et al.~\cite{suzuki1985} (\Cref{fig:table_mask}).
Smaller contours are filtered out, keeping only the two largest ones.
If the mask is not convex, additional internal contours might appear, which are also removed.
Using the probabilistic Hough transform~\cite{matas2000}, we identify and group similar lines to detect the table edges.
If exactly four lines remain, their intersections are calculated and sorted clockwise to determine the table’s four corners (\Cref{fig:table_corners}).
These corner points are then used for an initial camera calibration using the method described in \cref{ssec:calibration}. 
To lift the corner ordering ambiguity, we perform this calibration twice with different corner orderings and select the result with minimal reprojection error and a plausible transformation (e.g., the camera isn’t unrealistically far, typically under 10m).

This initial calibration can then be leveraged to detect other features such as the midline.
We detect the midline using a combination of Canny edge detection and the Hough line transform within the table mask as shown in \Cref{fig:table_midline}.
The detected lines are filtered based on the known directions of the table edges, and the midline points are defined by the intersection between the midline and the back edge of the table.
The midline points are not necessary for calibration but will improve the calibration accuracy.

%

\section{Trajectory Segmentation}

We use BlurBall~\cite{blurball} to detect the ball for a whole rally.
To analyze individual ball trajectories, we segment the rally using piecewise polynomial segmentation.
Since a ball's flight trajectory can be well approximated by a second-degree polynomial~\cite{wang2023e}, this approach naturally aligns with the motion characteristics of the ball.
These trajectory segments can be effectively identified through segmentation and solved efficiently using dynamic programming, as described in~\cite{duan2021}.
The solution is one that minimizes
\begin{equation}
    J = \sum^T_t \left( |u_t - P_u(t)| + |v_t - P_v(t)|\right) + \lambda K,
\end{equation}
where $(u_t, v_t)$ are the ball pixel coordinates at time $t$, $P_u$ and $P_v$ are respectively the piecewise polynomes fitted to $u$ and $v$, $\lambda$ is a penalty for creating new polynomes and $K$ is the number of segments in the piecewise polynomes.

This approach was selected for its robustness, as it is independent of the \ac{POV} and, unlike previous methods using GRU networks~\cite{liu2022b, ertner2024}, does not rely on the players' pose.
However, we observed that the number of frames between the table bounce and racket strike can be quite low.
Attacks require the ball to be hit as early as possible which then leads to obstruction by the player.
The number of balls observed in between can be too small to be detected as a new segment.
To solve this, we leverage the additional blur information provided by BlurBall~\cite{blurball}. 
The blur is linked to the ball's velocity and thus to the derivative of the polynomials.
\begin{equation}
    B_{error} =  \left| \theta_t - \arctan(\frac{P_v(t)^\prime}{P_u(t)^\prime}) \right|
\end{equation}
where $\theta_t$ is the observed motion blur.
We add this blur error $B_{error}$ to the cost function $J$.

The piecewise polynomial segmentation offers greater generalizability compared to heuristic approaches, such as detecting changes in the ball's velocity, as used in ~\cite{calandre2021a, shen2016}.

\begin{figure}
    \centering
    \includegraphics[width=0.9\linewidth]{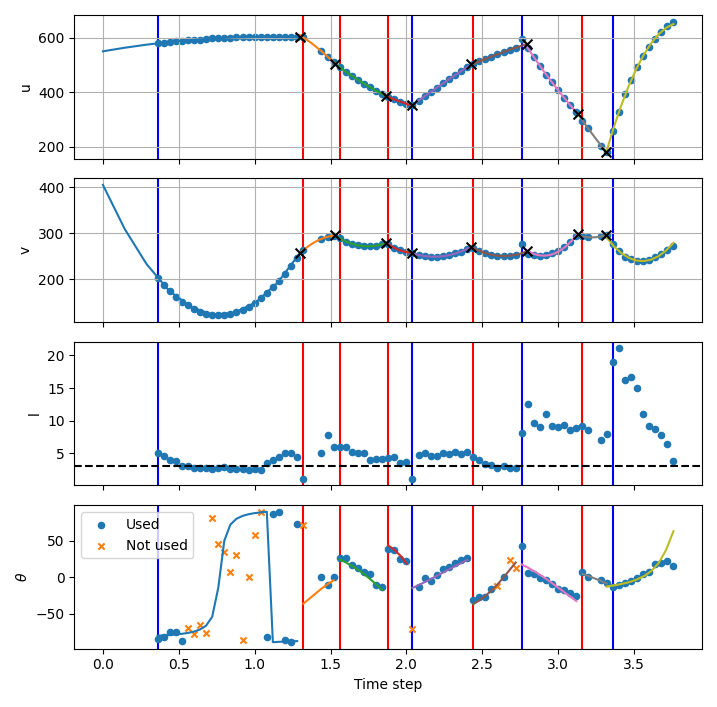}
    \caption{Segmentation of a rally. The dots represent the detected ball positions, while the vertical lines delineate different segments. Using heuristics, table bounces are marked in red and racket bounces in blue. The black crosses indicate the precise bounce coordinates, determined by the intersection of the fitted second-degree polynomials.}
    \label{fig:traj_seg}
\end{figure}

Once these polynomials are determined, bounce positions and times can be precisely calculated at the intersection points of consecutive polynomial segments.
This is extremely important as most recordings run at 25 fps. 
This would lead to an uncertainty in the bounce time of 40 ms (time between successive frames for 25 fps) and a bounce position error that can go up to 40 cm for a ball with a  velocity of 10 m/s.
As explained in \Cref{ssec:reconstruction}, we require a very accurate bounce position for our reconstruction to work.
This method offers higher accuracy than previous approaches, which are inherently limited by the video framerate.
An example of the segmentation is shown in \Cref{fig:traj_seg}.

To differentiate between racket strikes and table bounces, we employ a heuristic approach.
Specifically, we assume that a change in direction along the table's longitudinal axis indicates a racket strike.

\subsection{Bouncing Position}
\label{ssec:bounce_pos}
Throughout the ball's trajectory, there exists a discrete moment where its exact 3D position can be determined: the bounce.
At this instant, the ball is in direct contact with the table, whose pose is known from the camera calibration.
This constraint restricts the possible ball positions to a 3D plane defined by its normal vector $\bm{n}$ and a point $\bm{X_{plane}}$.
Furthermore, trajectory segmentation provides the precise image coordinates of the ball at the bounce point, allowing us to compute the corresponding ray direction as
\begin{equation}
\bm{d} = \bm{K}^{-1} 
\begin{bmatrix} u \\
v\\
1
\end{bmatrix} . 
\end{equation}
The intersection of this ray with the plane yields the 3D bounce position of the ball
\begin{equation} 
\bm{X_{bounce}} = \frac{\bm{X_{plane}} \cdot \bm{n}}{\bm{d} \cdot \bm{n}} \bm{d}. 
\end{equation}
However, the computed intersection lies on the court plane, whereas the actual contact point is slightly offset due to the ball’s radius.
This offset is particularly significant when the camera's viewpoint is low and close to the court plane, where perspective effects amplify small vertical errors.
We deal with this by shifting $\bm{X_{plane}}$ upwards by the radius of the ball with regard to the table. 
\section{3D Reconstruction}
\label{sec:3d_reconstruction}

Using the camera calibration and the segmented 2D ball trajectory, we reconstruct the 3D ball trajectory.
To achieve this, we leverage the ball dynamics described in \Cref{ssec:physics}.
These dynamics are used to optimize the ball's bounce velocity and spin, minimizing the reprojection error as detailed in \Cref{ssec:reconstruction}.

\subsection{Physics Model}
\label{ssec:physics} 
The ball dynamics are divided into two categories: the airborne motion of the ball and its behavior during bounces.
The airborne ball trajectory is defined by the following \ac{ODE}~\cite{wang2019}:
\begin{equation}
    m \dot{\bm{v}} = \underbrace{k_D ||\bm{v}|| \bm{v}}_\text{Drag} + \underbrace{k_M \bm{\omega} \times \bm{v}}_\text{Magnus} + m\bm{g}
    \label{eq:ode_aero}
\end{equation}
where $m= \SI{2.7e-3}{\kilogram}$ is the mass of the ball, $\bm{v}$ and $\bm{\omega}$ are respectively the ball velocity and spin, $k_D = \SI{3.8e-4}{\kilogram\cdot\second^{-4}}$ is the drag coefficient, $k_M=\SI{4.86e-6}{\kilogram\cdot\second^{-4}\cdot\metre^{-1}}$ is the Magnus coefficient and $\bm{g}$ is the gravity vector.
The \ac{ODE} in \Cref{eq:ode_aero} lacks an analytical solution due to the quadratic drag term and the Magnus force's dependence on spin and velocity.
However, it can be solved as an initial value problem using Runge-Kutta or collocation methods.

The bounce is a discrete event that is not described by an \ac{ODE}.
It is modeled using a Coulomb friction model~\cite{bao2012a} with:
\begin{equation}
\begin{split}
\bm{v^{+}} & = \bm{A} \bm{v^{-}} + \bm{B} \bm{\omega^{-}}\\
\bm{\omega^{+}} & = \bm{C} \bm{v^{-}} + \bm{D} \bm{\omega^{-}}.
\end{split}
\label{eqn:table_bounce}
\end{equation}
where $\bm{v^{-}}$, $\bm{\omega^{-}}$ and $\bm{v^{+}}$, $\bm{\omega^{+}}$ are respectively the ball velocity and spin before and after the bounce.
The dynamic matrices $\bm{A}$, $\bm{B}$, $\bm{C}$, and $\bm{D}$ will be different depending on whether the ball is rolling or sliding.
The bounce type can be distinguished with the coefficient $\alpha$.
\begin{equation}
\alpha = \frac{\mu\left(1+k_{COR}\right)\left|v^{-}_{z}\right|}{\sqrt{\left(v^{-}_{x}-\omega^{-}_{y} r\right)^2+\left(v^{-}_{y}+\omega_{x} r\right)^2}}
\end{equation}
where $k_{COR}=0.85$ is the \ac{COR}, $\mu=0.3$ is the friction coefficient and $r=\SI{0.02}{\meter}$ is the radius of the ball.

If $\alpha \geq 0.4$, then the ball is rolling and we use the following matrices:
\begin{equation}
\begin{aligned}
&\bm{A}=\left[\begin{array}{ccc}
0.6 & 0 & 0 \\
0 & 0.6 & 0 \\
0 & 0 & -k_{COR}
\end{array}\right] \quad \bm{B}=\left[\begin{array}{ccc}
0 & 0.4 r & 0 \\
-0.4 r & 0 & 0 \\
0 & 0 & 0
\end{array}\right] \\
& \bm{C}=\left[\begin{array}{ccc}
0 & -0.6 / r & 0 \\
0.6 / r & 0 & 0 \\
0 & 0 & 0
\end{array}\right] \quad \bm{D}=\left[\begin{array}{ccc}
0.4 & 0 & 0 \\
0 & 0.4 & 0 \\
0 & 0 & 1
\end{array}\right] .
\end{aligned}
\end{equation}
If $\alpha < 0.4$, then the ball is sliding and we use the following matrices:
\begin{equation}
\begin{aligned}
& \bm{A}=\left[\begin{array}{ccc}
1-\alpha & 0 & 0 \\
0 & 1-\alpha & 0 \\
0 & 0 & -k_{COR}
\end{array}\right] \quad \bm{B}=\left[\begin{array}{ccc}
0 & \alpha r & 0 \\
-\alpha r & 0 & 0 \\
0 & 0 & 0
\end{array}\right] \\
& \bm{C}=\left[\begin{array}{ccc}
0 & -\frac{3 \alpha}{2 r} & 0 \\
\frac{3 \alpha}{2 r} & 0 & 0 \\
0 & 0 & 0
\end{array}\right] \quad \bm{D}=\left[\begin{array}{ccc}
1-\frac{3}{2} \alpha & 0 & 0 \\
0 & 1-\frac{3}{2} \alpha & 0 \\
0 & 0 & 1
\end{array}\right] .
\end{aligned}
\end{equation}
%
%
Wang et al.~\cite{wang2023e} showed that, though the ball spin can be approximated by the curve in the trajectory due to the Mangus effect, it can be further refined from the change in direction after the bounce leading to accurate spin estimation.
Though we do not have access to the 3D ball trajectory, we can observe the ball's curve and the change in direction after the bounce.
Since we take into account the spin in the physical model, we are able to infer the spin of the ball from only the 2D ball trajectory.
Pre-existing methods that infer spin from monocular recordings usually rely on the player stroke movement~\cite{Sato2020ceur, Kulkarni2021cvpr}.

\subsection{Reconstruction}
\label{ssec:reconstruction}
From the trajectory segmentation, we extracted the exact image position and timestamp of the ball's bounce on the table.  
By computing the intersection of the image ray corresponding to the bounce position with the 3D table plane, we determine the precise 3D bounce position of the ball, $\bm{p}_0$.  
Unlike previous methods~\cite{shen2016, liu2022b, ertner2024}, which initialize their models at the start of the trajectory, we use the bounce as the initial state.  
This approach eliminates the need to optimize for the ball's position and instead focuses on optimizing its velocity and spin immediately before the bounce, represented as $\bm{v}^-$ and $\bm{\omega}^-$.  
Using a bounce model, we compute the ball's velocity $\bm{v}^+$  and spin $\bm{\omega}^+$ just after the bounce.  
We then estimate the 3D positions of the ball before and after the bounce, denoted as $\bm{\hat{X_k}}$.  
By optimizing $[\bm{v^-},\bm{\omega^-}]$ for the reprojection error, we reconstruct the 3D trajectory of the ball.  
We use IPopt~\cite{Wchter2006OnTI} in Casadi~\cite{Andersson2018} to optimize our loss function, defined as:  
\begin{equation}
    \mathcal{L}_{ball} = \sum_{k=0}^{n} ||\bm{x_k} - \hat{\bm{x_k}}||^2_2 = \sum_{k=0}^{n} ||\bm{x_k} - \bm{P} \hat{\bm{X_k}}||^2_2
\end{equation}  
where $\bm{x_k}$ represents the observed ball positions.  
The ball spin is initialized to $\bm{\omega^-} =[0, 0, 0]^T$ (rad/s) and the ball velocity to $\bm{v^-} = [0, \pm 5, -3]^T$ (m/s).



\section{3D Pose Estimation}
\label{sec:3d_pose}

So far, our primary focus has been on reconstructing the ball’s 3D trajectory.
While valuable on its own, this information becomes significantly more insightful when combined with player motion and stroke estimation.
By linking the ball's trajectory to the player's movements and actions, we can gain a deeper understanding of gameplay dynamics, strategy, and shot execution.  

To achieve this, we first track the players and estimate their 2D pose using RTMPose~\cite{jiang2023rtmpose}. 
We then infer their 3D pose with MotionBERT~\cite{motionbert2022}.
However, since the estimated 3D pose is expressed in camera coordinates, it must be transformed into the world frame.
Camera calibration provides the necessary rotation, but estimating the translation vector $\bm{T}$ and scale factor $s$ remains a challenge.  

\begin{figure}
    \centering
    \includegraphics[width=\linewidth]{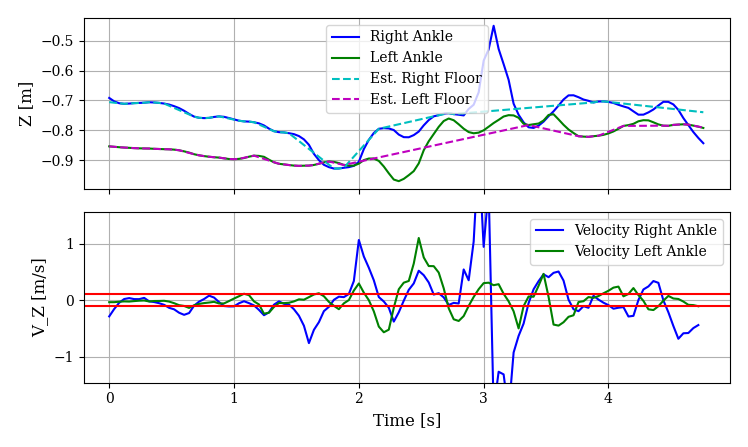}
    \caption{Floor contact estimation based on the vertical velocity of the players' ankles.}
    \label{fig:contact}
\end{figure}

We estimate \( \bm{T} \) by minimizing the reprojection error between the detected 2D keypoints and the projected 3D pose in the world frame:  
\begin{equation}
 \mathcal{L}_{\text{pose}} = \sum_{k=1}^{17} \left\| \bm{q}_k - s \bm{K} (\bm{Q}_k + \bm{T}) \right\|^2_2 + \lambda \mathcal{L}_{\text{floor}}
\end{equation}
where \( \bm{q}_k \) represents the observed 2D joint positions, $\bm{Q}_k$ denotes the inferred 3D joint positions in the camera frame, and $s$ is the scale factor of the projection and $\lambda=10$ is a weight coefficient.
$\mathcal{L}_{floor}$ is an additional constraint to ensure that the player's feet touch the ground and that the depth information is correctly estimated.  
Indeed, although the 3D pose model accurately captures lateral motion, it struggles with depth estimation relative to the camera.
This issue is particularly pronounced in side-view recordings, where precise depth perception is critical for reliable player tracking.  
%
%
We thus define $\mathcal{L}_{floor}$ as 
\begin{equation}
    \mathcal{L}_{\text{floor}} = \mathds{1}_{\text{contact}}(t) \left(\left\| \bm{q}_{\text{left ankle}} - 0.1 \right\|^2_2 + \left\| \bm{q}_{\text{right ankle}} - 0.1 \right\|^2_2 \right),
\end{equation}
where we approximate the ankle height as 0.1 m when the feet are touching the floor.
We define floor contact as periods when the vertical velocity of the ankles remains below a predefined threshold, indicating contact.
By constraining translation estimation to these stable frames, we obtain more accurate and consistent pose estimates.
Jumps would otherwise cause the estimated pose to shift backward. 
For frames where the feet are not in contact with the floor, we interpolate the translation linearly to ensure smooth motion across the sequence.
An example of floor contact estimation is shown in \Cref{fig:contact}.  

\section{Experiments}

While qualitative assessment of our pipeline can be performed through visual inspection of the reconstructions, obtaining a quantitative evaluation is considerably more challenging due to the lack of video recordings with calibrated cameras, synchronized human motion capture and 3D ball tracking.  
Therefore, we evaluate the individual components of our pipeline to the best of our ability.

\subsection{Camera Calibration}

We first evaluated the accuracy of table corner detection.
For this, we manually labeled the exact positions of the table corners in 40 images distinct from the training set
We achieved a \ac{MAE} of 2.39 $\pm$ 1.47 px.
%

\begin{figure}
    \centering
    \includegraphics[width=\linewidth]{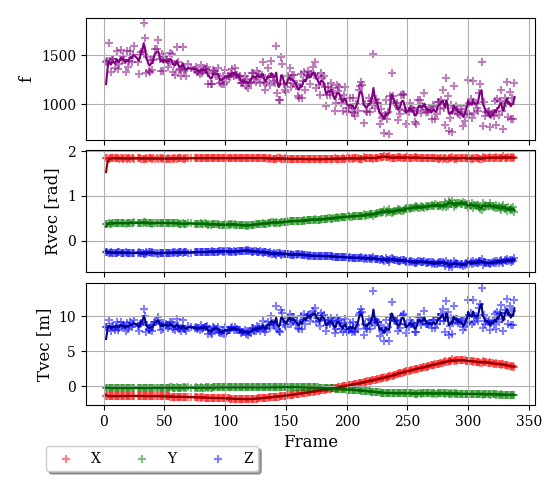}
    \caption{Continuous camera calibration from an online video. The darker lines are the filtered observations using a Kalman filter.}
    \label{fig:camera_tracking}
\end{figure}

We performed camera calibration on a video with camera movement and zooming~\footnote{First rally of \url{https://www.youtube.com/watch?v=nd40lIYtQmA}}.
The results, shown in \Cref{fig:camera_tracking}, demonstrate that the estimated parameters are temporally consistent.
Additionally, slight variations in $f$ closely align with changes in the Z component of $\bm{T}$, as their estimations are interdependent.

In the supplementary material, we also include multiple examples of successful camera calibration as well as examples of failure cases, which we go into more detail in \Cref{ssec:limitations}.

\subsection{3D Ball Trajectory Reconstruction Benchmark}
We recorded a table tennis match between two club-level players using a multi-camera system capable of tracking the ball's 3D position at 200 Hz.
However, the system requires orange balls, which BlurBall was not trained to detect, rendering the ball detector inapplicable.
Despite this limitation, the recorded 3D trajectories are accurate and can be used to generate simulated 2D observations at 25 fps using the pinhole camera model.
To achieve this, we created 130 ball trajectories from side, oblique, and back views, introducing noise to the 2D ball position ($\sigma_{p_b} = 2$ px) and to the blur estimation ($\sigma_\theta = 6^\circ$, $\sigma_l = 1$ px). 
The standard deviations were set according to the error observed in~\cite{blurball}.
The evaluation results are presented in \Cref{tab:evaluation}.
We can observe that our rally reconstruction is robust to noise. 
This is because, with enough observations, the constraints imposed by the physics filter out the noise.
While we can't directly validate the estimated bounce spin due to lack of ground truth, the low reconstruction error suggests it is reasonably accurate—since spinless trajectories would differ significantly, and topspin was applied during recording as instructed.
\begin{table}
    \centering
    \caption{3D Reconstruction errors using our method}
    \label{tab:evaluation}
    \resizebox{\columnwidth}{!}{%
    \begin{tabular}{lcc|cc}
        \toprule
        \textbf{POV} & \multicolumn{2}{c|}{\textbf{No Noise}} & \multicolumn{2}{c}{\textbf{Noise}} \\
        \cmidrule(lr){2-3} \cmidrule(lr){4-5}
         & \textbf{Succ. [\%]} & \textbf{MAE [cm]} & \textbf{Succ. [\%]} & \textbf{MAE [cm]} \\
        \midrule
        Side    & 97.3 & 8.9  & 97.3 & 12.4  \\
        Oblique & 91.5 & 14.5 & 89.9 & 17.1  \\
        Back    & 92.3 & 22.3 & 86.9 & 29.8  \\
        \bottomrule
    \end{tabular}%
    }
\end{table}

\subsection{Limitations}
\label{ssec:limitations}
While our camera calibration method is effective in most cases, it has certain limitations.
Failures primarily occur due to player-induced occlusions, where body contours may be misinterpreted as table edges.
This issue is particularly pronounced in back-view recordings, where the player frequently obstructs a significant portion of the table.
However, since players move dynamically throughout a rally, clear views of the table naturally emerge, providing reliable opportunities for calibration.
Additionally, if the camera's elevation angle is too low, the calibration process may become less reliable.
In such cases, the segmentation model doesn't perform as well because of the thin shape of the mask and the intersection of detected table edges becomes highly sensitive to minor errors, increasing the likelihood of calibration failure.

While the ball detection method remains effective with moving cameras, the trajectory segmentation does not, as the ball's observed motion no longer follows a polynomial trajectory.
Furthermore, our segmentation approach relies on second-degree polynomials, requiring the ball to be visible for at least three frames after a bounce.
In some cases, this condition is not met due to occlusions caused by players or when the ball is hit back early after bouncing, limiting the effectiveness of the trajectory segmentation process.

\section{Conclusion}

We developed a method capable of reconstructing the 3D trajectory of a table tennis ball from an online recording.
Moreover, the camera calibration is performed automatically compared to previous approaches.
With our approach, we can make use of the vast amount of table tennis match recordings available on the internet.
This can help get better player statistics or build a table tennis foundation model that could for example predict the next likely stroke or real-time win probability.
While our method has been tested exclusively on table tennis, it could be easily adapted for other racket sports like tennis or pickleball.

\newpage

\section*{ACKNOWLEDGMENT}
Thank you to Dieter Buechler and Jan Schneider for their help with the recording setup and to Giorgio Becherini and Chuyu Yang for participating in the recordings.

{
    \small
    \bibliographystyle{ieeenat_fullname}
    \bibliography{main}

\begin{thebibliography}{66}
\providecommand{\natexlab}[1]{#1}
\providecommand{\url}[1]{\texttt{#1}}
\expandafter\ifx\csname urlstyle\endcsname\relax
  \providecommand{\doi}[1]{doi: #1}\else
  \providecommand{\doi}{doi: \begingroup \urlstyle{rm}\Url}\fi

\bibitem[zot()]{zotero-2360}
How {{To Experience More Time In Tennis}} {\textbar} {{Feel Tennis}}.
\newblock https://www.feeltennis.net/experience-more-time/.

\bibitem[A(2017)]{a2017}
J. A.
\newblock The fastest sport? {{Table Tennis VS Badminton}}, 2017.

\bibitem[Andersson et~al.(2018)Andersson, Gillis, Horn, Rawlings, and Diehl]{Andersson2018}
Joel A~E Andersson, Joris Gillis, Greg Horn, James~B Rawlings, and Moritz Diehl.
\newblock {CasADi} -- {A} software framework for nonlinear optimization and optimal control.
\newblock \emph{Mathematical Programming Computation}, 2018.

\bibitem[Anonymous(2025)]{blurball}
Anonymous.
\newblock Blurball: ball detection with blur estimation, 2025.

\bibitem[Archana and Geetha(2015)]{archana2015}
M. Archana and M.~Kalaisevi Geetha.
\newblock Object {{Detection}} and {{Tracking Based}} on {{Trajectory}} in {{Broadcast Tennis Video}}.
\newblock \emph{Procedia Computer Science}, 58:\penalty0 225--232, 2015.

\bibitem[Bao et~al.(2012)Bao, Chen, Wang, Pan, and Meng]{bao2012a}
H. Bao, X. Chen, Z. Wang, M. Pan, and F. Meng.
\newblock Bouncing model for the table tennis trajectory prediction and the strategy of hitting the ball.
\newblock In \emph{2012 {{IEEE International Conference}} on {{Mechatronics}} and {{Automation}}}, pages 2002--2006, 2012.

\bibitem[Calandre et~al.(2021)Calandre, P{\'e}teri, Mascarilla, and Tremblais]{calandre2021a}
J. Calandre, R. P{\'e}teri, L. Mascarilla, and B. Tremblais.
\newblock Extraction and analysis of {{3D}} kinematic parameters of {{Table Tennis}} ball from a single camera.
\newblock In \emph{2020 25th {{International Conference}} on {{Pattern Recognition}} ({{ICPR}})}, pages 9468--9475, Milan, Italy, 2021. IEEE.

\bibitem[Chaurasia and Culurciello(2017)]{linknet}
Abhishek Chaurasia and Eugenio Culurciello.
\newblock Linknet: Exploiting encoder representations for efficient semantic segmentation.
\newblock In \emph{2017 IEEE Visual Communications and Image Processing (VCIP)}, pages 1--4, 2017.

\bibitem[Chen et~al.(2009)Chen, Tien, {Yi-Wen Chen}, Chen, Tsai, and Lee]{chen2009}
Hua-Tsung Chen, Ming-Chun Tien, {Yi-Wen Chen}, Yi-Wen Chen, Wen-Jiin Tsai, and Suh-Yin Lee.
\newblock Physics-based ball tracking and {{3D}} trajectory reconstruction with applications to shooting location estimation in basketball video.
\newblock \emph{Journal of Visual Communication and Image Representation}, 20\penalty0 (3):\penalty0 204--216, 2009.

\bibitem[Chen et~al.(2012)Chen, Tsai, Tsai, Lee, and Yu]{chen2012}
Hua-Tsung Chen, Wen-Jiin Tsai, Wen-Jiin Tsai, Suh-Yin Lee, and Jen-Yu Yu.
\newblock Ball tracking and {{3D}} trajectory approximation with applications to tactics analysis from single-camera volleyball sequences.
\newblock \emph{Multimedia Tools and Applications}, 60\penalty0 (3):\penalty0 641--667, 2012.

\bibitem[Chen et~al.(2017)Chen, Papandreou, Schroff, and Adam]{deeplabv3}
Liang-Chieh Chen, George Papandreou, Florian Schroff, and Hartwig Adam.
\newblock Rethinking atrous convolution for semantic image segmentation, 2017.

\bibitem[Chen et~al.(2018)Chen, Zhu, Papandreou, Schroff, and Adam]{deeplabv3plus}
Liang-Chieh Chen, Yukun Zhu, George Papandreou, Florian Schroff, and Hartwig Adam.
\newblock Encoder-decoder with atrous separable convolution for semantic image segmentation.
\newblock In \emph{Computer Vision -- ECCV 2018}, pages 833--851, Cham, 2018. Springer International Publishing.

\bibitem[Chen and Wang(2024)]{chen2024}
Yu-Jou Chen and Yu-Shuen Wang.
\newblock Tracknetv3: Enhancing shuttlecock tracking with augmentations and trajectory rectification.
\newblock In \emph{Proceedings of the 5th ACM International Conference on Multimedia in Asia}, New York, NY, USA, 2024. Association for Computing Machinery.

\bibitem[Duan et~al.(2021)Duan, Wang, and Wang]{duan2021}
J. Duan, Q. Wang, and Y. Wang.
\newblock {{HOPS}}: {{A Fast Algorithm}} for {{Segmenting Piecewise Polynomials}} of {{Arbitrary Orders}}.
\newblock \emph{IEEE Access}, 9:\penalty0 155977--155987, 2021.

\bibitem[Ertner et~al.(2024)Ertner, Konglevoll, Ibh, and Gra{\ss}hof]{ertner2024}
M.H. Ertner, S.~S. Konglevoll, M. Ibh, and S. Gra{\ss}hof.
\newblock {{SynthNet}}: {{Leveraging Synthetic Data}} for {{3D Trajectory Estimation}} from {{Monocular Video}}.
\newblock In \emph{Proceedings of the 7th {{ACM International Workshop}} on {{Multimedia Content Analysis}} in {{Sports}}}, pages 51--58, New York, NY, USA, 2024. Association for Computing Machinery.

\bibitem[Farin et~al.(2004)Farin, Krabbe, With, and Effelsberg]{farin2004a}
Dirk Farin, Susanne Krabbe, Peter With, and Wolfgang Effelsberg.
\newblock \emph{Robust Camera Calibration for Sport Videos Using Court Models}.
\newblock 2004.

\bibitem[Han et~al.(2006)Han, Farin, and de~With]{han2006}
Jungong Han, Dirk Farin, and Peter H.~N. de With.
\newblock Generic 3-d modeling for content analysis of court-net sports sequences.
\newblock In \emph{Advances in Multimedia Modeling}, pages 279--288, Berlin, Heidelberg, 2006. Springer Berlin Heidelberg.

\bibitem[Huang et~al.(2019)Huang, Liao, Chen, {\.I}k, and Peng]{huang2019}
Y. Huang, I. Liao, C. Chen, T. {\.I}k, and W. Peng.
\newblock {{TrackNet}}: {{A Deep Learning Network}} for {{Tracking High-speed}} and {{Tiny Objects}} in {{Sports Applications}}.
\newblock In \emph{2019 16th {{IEEE International Conference}} on {{Advanced Video}} and {{Signal Based Surveillance}} ({{AVSS}})}, pages 1--8, 2019.

\bibitem[Hung(2018)]{hung2018}
C. Hung.
\newblock A {{Study}} of {{Automatic}} and {{Real-Time Table Tennis Fault Serve Detection System}}.
\newblock \emph{Sports}, 6\penalty0 (4):\penalty0 158, 2018.

\bibitem[Iakubovskii(2019)]{Iakubovskii:2019}
Pavel Iakubovskii.
\newblock Segmentation models pytorch.
\newblock \url{https://github.com/qubvel/segmentation_models.pytorch}, 2019.

\bibitem[Jiang et~al.(2023)Jiang, Lu, Zhang, Ma, Han, Lyu, Li, and Chen]{jiang2023rtmpose}
Tao Jiang, Peng Lu, Li Zhang, Ningsheng Ma, Rui Han, Chengqi Lyu, Yining Li, and Kai Chen.
\newblock Rtmpose: Real-time multi-person pose estimation based on mmpose.
\newblock \emph{arXiv preprint arXiv:2303.07399}, 2023.

\bibitem[Kamble et~al.(2019)Kamble, {Avinash G. Keskar}, Keskar, and Bhurchandi]{kamble2019}
Paresh~R. Kamble, {Avinash G. Keskar}, Avinash~G. Keskar, and Kishor~M. Bhurchandi.
\newblock Ball tracking in sports: A survey.
\newblock \emph{Artificial Intelligence Review}, 52\penalty0 (3):\penalty0 1655--1705, 2019.

\bibitem[Kanaeva et~al.(2015)Kanaeva, Gurevich, and Vakhitov]{kanaeva2015}
Ekaterina Kanaeva, Lev Gurevich, and Alexander Vakhitov.
\newblock Camera pose and focal length estimation using regularized distance constraints.
\newblock In \emph{BMVC}, pages 162--1, 2015.

\bibitem[Kirillov et~al.(2023)Kirillov, Mintun, Ravi, Mao, Rolland, Gustafson, Xiao, Whitehead, Berg, Lo, Doll{\'a}r, and Girshick]{kirillov2023segany}
Alexander Kirillov, Eric Mintun, Nikhila Ravi, Hanzi Mao, Chloe Rolland, Laura Gustafson, Tete Xiao, Spencer Whitehead, Alexander~C. Berg, Wan-Yen Lo, Piotr Doll{\'a}r, and Ross Girshick.
\newblock Segment anything.
\newblock \emph{arXiv:2304.02643}, 2023.

\bibitem[Kulkarni et~al.(2022)Kulkarni, Jamadagni, Paul, and Shenoy]{kulkarni2022}
K. Kulkarni, R. Jamadagni, J. Paul, and S. Shenoy.
\newblock Table {{Tennis Stroke Detection}} and {{Recognition Using Ball Trajectory Data}}.
\newblock \emph{SSRN Electronic Journal}, 2022.

\bibitem[Kulkarni and Shenoy(2021)]{Kulkarni2021cvpr}
Kaustubh~Milind Kulkarni and Sucheth Shenoy.
\newblock Table tennis stroke recognition using two-dimensional human pose estimation.
\newblock In \emph{Proceedings of the IEEE/CVF Conference on Computer Vision and Pattern Recognition (CVPR) Workshops}, pages 4576--4584, 2021.

\bibitem[Laffaye et~al.(2015)Laffaye, Phomsoupha, and Dor]{laffaye2015}
Guillaume Laffaye, Michael Phomsoupha, and Fr{\'e}d{\'e}ric Dor.
\newblock Changes in the {{Game Characteristics}} of a {{Badminton Match}}: {{A Longitudinal Study}} through the {{Olympic Game Finals Analysis}} in {{Men}}'s {{Singles}}.
\newblock \emph{Journal of Sports Science \& Medicine}, 14\penalty0 (3):\penalty0 584--590, 2015.

\bibitem[Li et~al.(2018)Li, Xiong, An, and Wang]{pan}
Hanchao Li, Pengfei Xiong, Jie An, and Lingxue Wang.
\newblock Pyramid attention network for semantic segmentation, 2018.

\bibitem[Li et~al.(2022)Li, Zheng, Zhang, Duan, Su, Wang, and Atkinson]{manet}
Rui Li, Shunyi Zheng, Ce Zhang, Chenxi Duan, Jianlin Su, Libo Wang, and Peter~M. Atkinson.
\newblock Multiattention network for semantic segmentation of fine-resolution remote sensing images.
\newblock \emph{IEEE Transactions on Geoscience and Remote Sensing}, 60:\penalty0 1–13, 2022.

\bibitem[Lin et~al.(2017)Lin, Dollár, Girshick, He, Hariharan, and Belongie]{fpn}
Tsung-Yi Lin, Piotr Dollár, Ross Girshick, Kaiming He, Bharath Hariharan, and Serge Belongie.
\newblock Feature pyramid networks for object detection.
\newblock In \emph{2017 IEEE Conference on Computer Vision and Pattern Recognition (CVPR)}, pages 936--944, 2017.

\bibitem[Liu and Wang(2022)]{liu2022b}
P. Liu and J. Wang.
\newblock {{MonoTrack}}: {{Shuttle}} trajectory reconstruction from monocular badminton video.
\newblock In \emph{2022 {{IEEE}}/{{CVF Conference}} on {{Computer Vision}} and {{Pattern Recognition Workshops}} ({{CVPRW}})}, pages 3512--3521, New Orleans, LA, USA, 2022. IEEE.

\bibitem[Matas et~al.(2000)Matas, Galambos, and Kittler]{matas2000}
J. Matas, C. Galambos, and J. Kittler.
\newblock Robust {{Detection}} of {{Lines Using}} the {{Progressive Probabilistic Hough Transform}}.
\newblock \emph{Computer Vision and Image Understanding}, 78\penalty0 (1):\penalty0 119--137, 2000.

\bibitem[Metzler and Pagel(2013-05-20/2013-05-23)]{metzler2013}
Juergen Metzler and Frank Pagel.
\newblock {{3D Trajectory Reconstruction}} of the {{Soccer Ball}} for {{Single Static Camera Systems}}.
\newblock In \emph{International {{Conference}} on {{Machine Vision Applications}}}, Kyoto, Japan, 2013-05-20/2013-05-23.

\bibitem[Muelling et~al.(2014)Muelling, Boularias, Mohler, Sch{\"o}lkopf, and Peters]{muelling2014c}
Katharina Muelling, Abdeslam Boularias, Betty Mohler, Bernhard Sch{\"o}lkopf, and Jan Peters.
\newblock Learning strategies in table tennis using inverse reinforcement learning.
\newblock \emph{Biological Cybernetics}, 108\penalty0 (5):\penalty0 603--619, 2014.

\bibitem[Naik et~al.(2022)Naik, Hashmi, and Bokde]{naik2022}
B.~T. Naik, M.~F. Hashmi, and N.D. Bokde.
\newblock A {{Comprehensive Review}} of {{Computer Vision}} in {{Sports}}: {{Open Issues}}, {{Future Trends}} and {{Research Directions}}.
\newblock \emph{Applied Sciences}, 12\penalty0 (9):\penalty0 4429, 2022.

\bibitem[Olson(2011)]{apriltag}
Edwin Olson.
\newblock Apriltag: A robust and flexible visual fiducial system.
\newblock In \emph{2011 IEEE International Conference on Robotics and Automation}, pages 3400--3407, 2011.

\bibitem[Penate-Sanchez et~al.(2013)Penate-Sanchez, Andrade-Cetto, and Moreno-Noguer]{penatesanchez2013}
Adrian Penate-Sanchez, Juan Andrade-Cetto, and Francesc Moreno-Noguer.
\newblock Exhaustive linearization for robust camera pose and focal length estimation.
\newblock \emph{IEEE Transactions on Pattern Analysis and Machine Intelligence}, 35\penalty0 (10):\penalty0 2387--2400, 2013.

\bibitem[Radosavovic et~al.(2020)Radosavovic, Kosaraju, Girshick, He, and Doll{\'a}r]{radosavovic2020}
Ilija Radosavovic, Raj~Prateek Kosaraju, Ross Girshick, Kaiming He, and Piotr Doll{\'a}r.
\newblock Designing {{Network Design Spaces}}.
\newblock In \emph{2020 {{IEEE}}/{{CVF Conference}} on {{Computer Vision}} and {{Pattern Recognition}} ({{CVPR}})}, pages 10425--10433, 2020.

\bibitem[Raj et~al.(2024)Raj, Wang, and Gedeon]{raj2024}
Arjun Raj, Lei Wang, and Tom Gedeon.
\newblock {{TrackNetV4}}: {{Enhancing Fast Sports Object Tracking}} with {{Motion Attention Maps}}, 2024.

\bibitem[Ravi et~al.(2024)Ravi, Gabeur, Hu, Hu, Ryali, Ma, Khedr, Rädle, Rolland, Gustafson, Mintun, Pan, Alwala, Carion, Wu, Girshick, Dollár, and Feichtenhofer]{ravi2024sam2segmentimages}
Nikhila Ravi, Valentin Gabeur, Yuan-Ting Hu, Ronghang Hu, Chaitanya Ryali, Tengyu Ma, Haitham Khedr, Roman Rädle, Chloe Rolland, Laura Gustafson, Eric Mintun, Junting Pan, Kalyan~Vasudev Alwala, Nicolas Carion, Chao-Yuan Wu, Ross Girshick, Piotr Dollár, and Christoph Feichtenhofer.
\newblock Sam 2: Segment anything in images and videos, 2024.

\bibitem[Robinson and Robinson(2018)]{robinson2018}
G. Robinson and I. Robinson.
\newblock Model trajectories for a spinning tennis ball: I. the service stroke.
\newblock \emph{Physica Scripta}, 93, 2018.

\bibitem[Ronneberger et~al.(2015)Ronneberger, Fischer, and Brox]{Unet}
Olaf Ronneberger, Philipp Fischer, and Thomas Brox.
\newblock U-net: Convolutional networks for biomedical image segmentation.
\newblock In \emph{Medical Image Computing and Computer-Assisted Intervention -- MICCAI 2015}, pages 234--241, Cham, 2015. Springer International Publishing.

\bibitem[Sato and Aono(2020)]{Sato2020ceur}
Soichiro Sato and Masaki Aono.
\newblock Leveraging human pose estimation model for stroke classification in table tennis.
\newblock In \emph{Working Notes Proceedings of the MediaEval 2020 Workshop, Online, 14-15 December 2020}. CEUR-WS.org, 2020.

\bibitem[Shen et~al.(2016)Shen, Liu, Li, and Yue]{shen2016}
L. Shen, Q. Liu, L. Li, and H. Yue.
\newblock {{3D}} reconstruction of ball trajectory from a single camera in the ball game.
\newblock In \emph{Proceedings of the 10th International Symposium on Computer Science in Sports ({{ISCSS}})}, pages 33--39, Cham, 2016. Springer International Publishing.

\bibitem[Sudre et~al.(2017)Sudre, Li, Vercauteren, Ourselin, and Jorge~Cardoso]{dice}
Carole~H. Sudre, Wenqi Li, Tom Vercauteren, Sebastien Ourselin, and M. Jorge~Cardoso.
\newblock Generalised dice overlap as a deep learning loss function for highly unbalanced segmentations.
\newblock In \emph{Deep Learning in Medical Image Analysis and Multimodal Learning for Clinical Decision Support}, pages 240--248, Cham, 2017. Springer International Publishing.

\bibitem[Sun et~al.(2020)Sun, Lin, Chuang, Hsu, Yu, Chung, and {\.I}k]{sun2020}
N. Sun, Y. Lin, S. Chuang, T. Hsu, D. Yu, H. Chung, and T. {\.I}k.
\newblock {{TrackNetV2}}: {{Efficient Shuttlecock Tracking Network}}.
\newblock In \emph{2020 {{International Conference}} on {{Pervasive Artificial Intelligence}} ({{ICPAI}})}, pages 86--91, 2020.

\bibitem[Suzuki and {be}(1985)]{suzuki1985}
Satoshi Suzuki and KeiichiA {be}.
\newblock Topological structural analysis of digitized binary images by border following.
\newblock \emph{Computer Vision, Graphics, and Image Processing}, 30\penalty0 (1):\penalty0 32--46, 1985.

\bibitem[Tamaki and Saito(2013)]{tamaki2013}
S. Tamaki and H. Saito.
\newblock Reconstruction of {{3D Trajectories}} for {{Performance Analysis}} in {{Table Tennis}}.
\newblock In \emph{2013 {{IEEE Conference}} on {{Computer Vision}} and {{Pattern Recognition Workshops}}}, pages 1019--1026, 2013.

\bibitem[Tarashima et~al.(2023)Tarashima, Haq, Wang, and Tagawa]{tarashima2023}
S. Tarashima, M. Haq, Y. Wang, and N. Tagawa.
\newblock Widely applicable strong baseline for sports ball detection and tracking.
\newblock In \emph{34th British Machine Vision Conference 2023, {{BMVC}} 2023, Aberdeen, {{UK}}, November 20-24, 2023}. BMVA, 2023.

\bibitem[Tebbe et~al.(2019)Tebbe, Gao, Sastre-Rienietz, and Zell]{tebbe2019}
J. Tebbe, Y. Gao, M. Sastre-Rienietz, and A. Zell.
\newblock A table tennis robot system using an industrial kuka robot arm.
\newblock In \emph{Pattern Recognition: 40th German Conference, GCPR 2018, Stuttgart, Germany, October 9-12, 2018, Proceedings 40}, pages 33--45. Springer, 2019.

\bibitem[W{\"a}chter and Biegler(2006)]{Wchter2006OnTI}
Andreas W{\"a}chter and Lorenz~T. Biegler.
\newblock On the implementation of an interior-point filter line-search algorithm for large-scale nonlinear programming.
\newblock \emph{Mathematical Programming}, 106:\penalty0 25--57, 2006.

\bibitem[Wang et~al.(2019)Wang, Zhang, Jin, and Ru]{wang2019}
Ping Wang, Qian Zhang, Yinli Jin, and Feng Ru.
\newblock Studies and simulations on the flight trajectories of spinning table tennis ball via high-speed camera vision tracking system.
\newblock \emph{Proceedings of the Institution of Mechanical Engineers, Part P}, 233\penalty0 (2):\penalty0 210--226, 2019.

\bibitem[Wang et~al.(2023)Wang, Sun, Luo, Zhang, Zhang, Dong, He, Zhang, Cheng, and Song]{wang2023e}
Yuxin Wang, Zhiyong Sun, Yongle Luo, Haibo Zhang, Wen Zhang, Kun Dong, Qiyu He, Qiang Zhang, Erkang Cheng, and Bo Song.
\newblock A {{Novel Trajectory-Based Ball Spin Estimation Method}} for {{Table Tennis Robot}}.
\newblock \emph{IEEE Transactions on Industrial Electronics}, pages 1--11, 2023.

\bibitem[Wong et~al.(2023)Wong, Myint, Dooley, and Hopgood]{wong2023}
P. Wong, H. Myint, L. Dooley, and A. Hopgood.
\newblock A multi-view automatic table tennis umpiring framework.
\newblock \emph{Proceedings of the Institution of Mechanical Engineers, Part P: Journal of Sports Engineering and Technology}, page 175433712311714, 2023.

\bibitem[Wu(2015)]{wu2015}
Changchang Wu.
\newblock P3.{{5P}}: {{Pose}} estimation with unknown focal length.
\newblock In \emph{2015 {{IEEE Conference}} on {{Computer Vision}} and {{Pattern Recognition}} ({{CVPR}})}, pages 2440--2448, 2015.

\bibitem[Wu et~al.(2019)Wu, Perteneder, and Koike]{wu2019}
Erwin Wu, Florian Perteneder, and Hideki Koike.
\newblock Real-time {{Table Tennis Forecasting System}} based on {{Long Short-term Pose Prediction Network}}.
\newblock In \emph{{{SIGGRAPH Asia}} 2019 {{Posters}}}, pages 1--2, New York, NY, USA, 2019. Association for Computing Machinery.

\bibitem[{Xinguo Yu} et~al.(2009){Xinguo Yu}, Yu, {Nan Jiang}, Jiang, {Loong-Fah Cheong}, Cheong, {Hon Wai Leong}, Leong, {Xiaogang Yan}, and Yan]{xinguoyu2009}
{Xinguo Yu}, Xinguo Yu, {Nan Jiang}, Nianjuan Jiang, {Loong-Fah Cheong}, Loong-Fah Cheong, {Hon Wai Leong}, Wai Leong, {Xiaogang Yan}, and Xin Yan.
\newblock Automatic camera calibration of broadcast tennis video with applications to {{3D}} virtual content insertion and ball detection and tracking.
\newblock \emph{Computer Vision and Image Understanding}, 113\penalty0 (5):\penalty0 643--652, 2009.

\bibitem[Yang et~al.(2024)Yang, Kang, Huang, Zhao, Xu, Feng, and Zhao]{depth_anything_v2}
Lihe Yang, Bingyi Kang, Zilong Huang, Zhen Zhao, Xiaogang Xu, Jiashi Feng, and Hengshuang Zhao.
\newblock Depth anything v2.
\newblock \emph{arXiv:2406.09414}, 2024.

\bibitem[Zandycke and Vleeschouwer(2022)]{zandycke2022}
Gabriel~Van Zandycke and Christophe~De Vleeschouwer.
\newblock {{3D Ball Localization From A Single Calibrated Image}}.
\newblock In \emph{2022 {{IEEE}}/{{CVF Conference}} on {{Computer Vision}} and {{Pattern Recognition Workshops}} ({{CVPRW}})}, pages 3471--3479. IEEE Computer Society, 2022.

\bibitem[Zhang et~al.(2023)Zhang, Yuan, Makoviychuk, Guo, Fidler, Peng, and Fatahalian]{zhang2023vid2player3d}
Haotian Zhang, Ye Yuan, Viktor Makoviychuk, Yunrong Guo, Sanja Fidler, Xue~Bin Peng, and Kayvon Fatahalian.
\newblock Learning physically simulated tennis skills from broadcast videos.
\newblock \emph{ACM Trans. Graph.}, 2023.

\bibitem[Zhao et~al.(2017)Zhao, Shi, Qi, Wang, and Jia]{pspnet}
Hengshuang Zhao, Jianping Shi, Xiaojuan Qi, Xiaogang Wang, and Jiaya Jia.
\newblock Pyramid scene parsing network.
\newblock In \emph{2017 IEEE Conference on Computer Vision and Pattern Recognition (CVPR)}, pages 6230--6239, 2017.

\bibitem[Zheng et~al.(2023)Zheng, Wu, Chen, Yang, Zhu, Shen, Kehtarnavaz, and Shah]{zheng2023}
Ce Zheng, Wenhan Wu, Chen Chen, Taojiannan Yang, Sijie Zhu, Ju Shen, Nasser Kehtarnavaz, and Mubarak Shah.
\newblock Deep {{Learning-based Human Pose Estimation}}: {{A Survey}}.
\newblock \emph{ACM Comput. Surv.}, 56\penalty0 (1):\penalty0 11:1--11:37, 2023.

\bibitem[Zheng and Kneip(2016)]{zheng2016a}
Yinqiang Zheng and Laurent Kneip.
\newblock A {{Direct Least-Squares Solution}} to the {{PnP Problem}} with {{Unknown Focal Length}}.
\newblock In \emph{2016 {{IEEE Conference}} on {{Computer Vision}} and {{Pattern Recognition}} ({{CVPR}})}, pages 1790--1798, Las Vegas, NV, USA, 2016. IEEE.

\bibitem[Zheng et~al.(2014)Zheng, Sugimoto, Sato, and Okutomi]{zheng2014}
Yinqiang Zheng, Shigeki Sugimoto, Imari Sato, and Masatoshi Okutomi.
\newblock A {{General}} and {{Simple Method}} for {{Camera Pose}} and {{Focal Length Determination}}.
\newblock In \emph{2014 {{IEEE Conference}} on {{Computer Vision}} and {{Pattern Recognition}}}, pages 430--437, Columbus, OH, USA, 2014. IEEE.

\bibitem[Zhou et~al.(2018)Zhou, Rahman~Siddiquee, Tajbakhsh, and Liang]{zhou2018}
Zongwei Zhou, Md~Mahfuzur Rahman~Siddiquee, Nima Tajbakhsh, and Jianming Liang.
\newblock {{UNet}}++: {{A Nested U-Net Architecture}} for {{Medical Image Segmentation}}.
\newblock In \emph{Deep {{Learning}} in {{Medical Image Analysis}} and {{Multimodal Learning}} for {{Clinical Decision Support}}}, pages 3--11, Cham, 2018. Springer International Publishing.

\bibitem[Zhu et~al.(2023)Zhu, Ma, Liu, Liu, Wu, and Wang]{motionbert2022}
Wentao Zhu, Xiaoxuan Ma, Zhaoyang Liu, Libin Liu, Wayne Wu, and Yizhou Wang.
\newblock Motionbert: A unified perspective on learning human motion representations.
\newblock In \emph{Proceedings of the IEEE/CVF International Conference on Computer Vision}, 2023.

\end{thebibliography}
}

\clearpage
\setcounter{page}{1}
\maketitlesupplementary

\begin{figure*}
    \centering
    \begin{subfigure}{0.23\linewidth}
        \includegraphics[width=\linewidth]{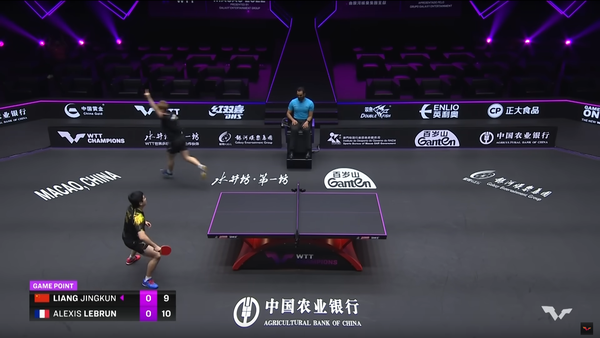}
    \end{subfigure}
    \begin{subfigure}{0.23\linewidth}
        \includegraphics[width=\linewidth]{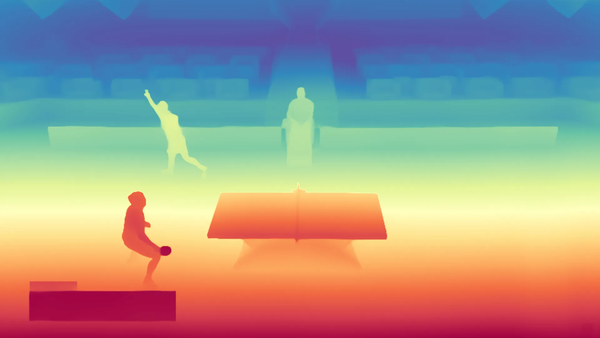}
    \end{subfigure}
    \begin{subfigure}{0.23\linewidth}
        \includegraphics[width=\linewidth]{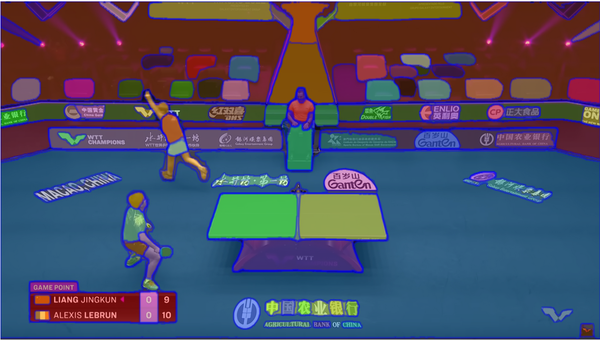}
    \end{subfigure}
    \begin{subfigure}{0.23\linewidth}
        \includegraphics[width=\linewidth]{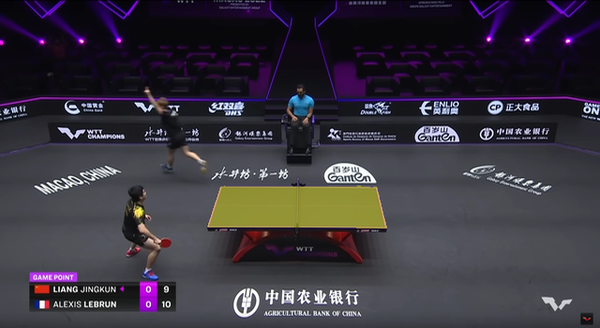}
    \end{subfigure}

    \begin{subfigure}{0.23\linewidth}
        \includegraphics[width=\linewidth]{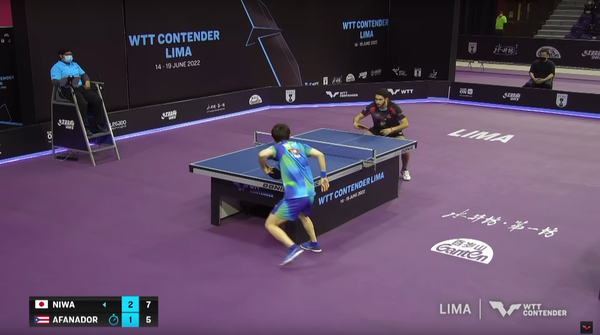}
    \end{subfigure}
    \begin{subfigure}{0.23\linewidth}
        \includegraphics[width=\linewidth]{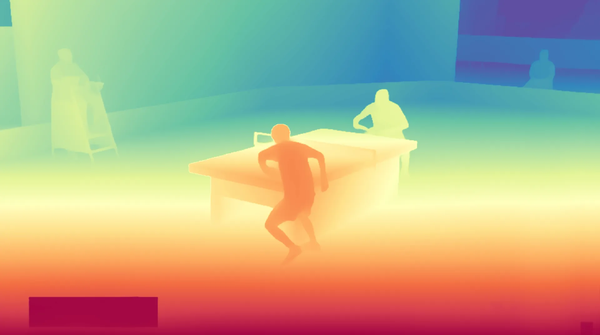}
    \end{subfigure}
    \begin{subfigure}{0.23\linewidth}
        \includegraphics[width=\linewidth]{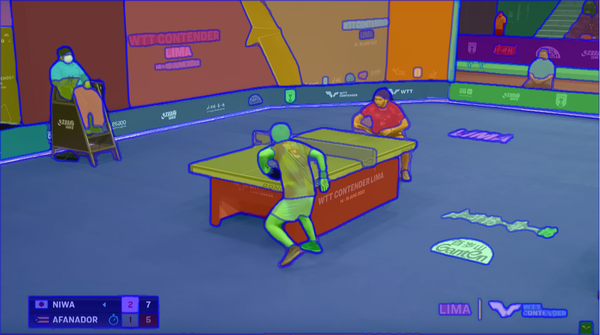}
    \end{subfigure}
    \begin{subfigure}{0.23\linewidth}
        \includegraphics[width=\linewidth]{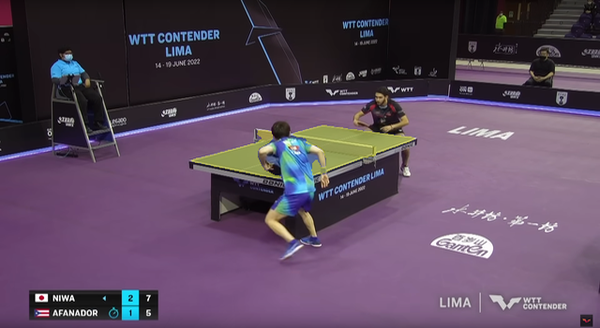}
    \end{subfigure}

    \begin{subfigure}{0.23\linewidth}
        \includegraphics[width=\linewidth]{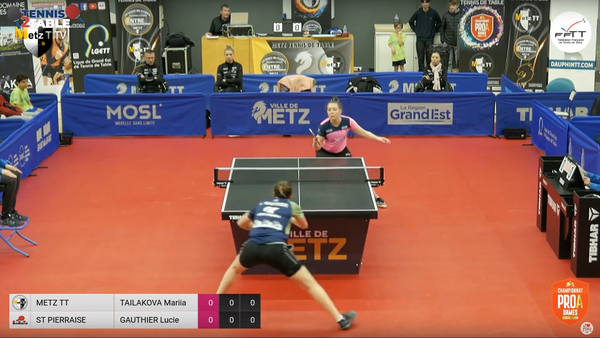}
    \end{subfigure}
    \begin{subfigure}{0.23\linewidth}
        \includegraphics[width=\linewidth]{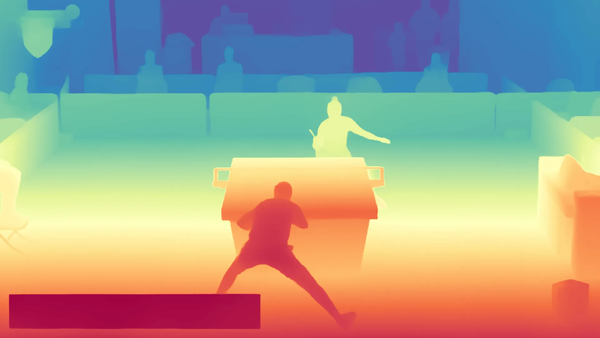}
    \end{subfigure}
    \begin{subfigure}{0.23\linewidth}
        \includegraphics[width=\linewidth]{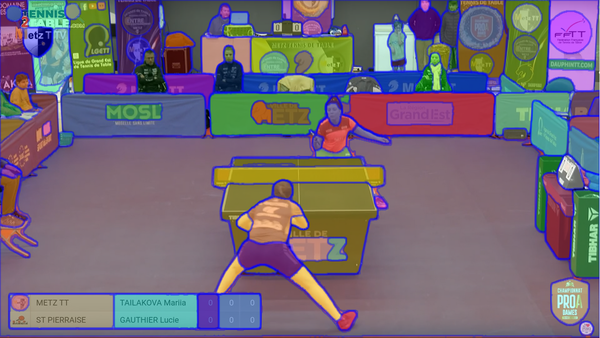}
    \end{subfigure}
    \begin{subfigure}{0.23\linewidth}
        \includegraphics[width=\linewidth]{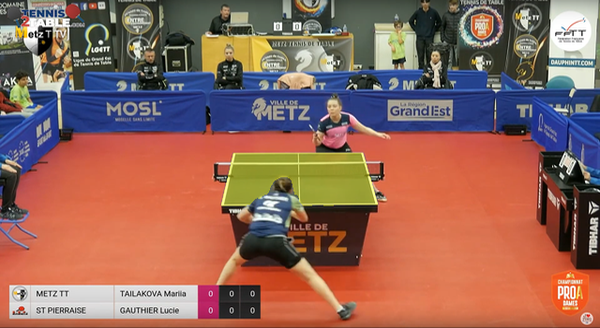}
    \end{subfigure}

    \caption{Comparing different segmentation methods for table tennis table. From left to right: Input image, Depth Anything V2~\cite{depth_anything_v2}, SAM~\cite{kirillov2023segany}, Custom trained Unet++ (ours)}
    \label{fig:comparing_table_masks}
\end{figure*}

\section{Table Segmentation}

Off-the-shelf segmentation models such as SAM~\cite{kirillov2023segany} and monocular depth estimation models like Depth Anything V2~\cite{depth_anything_v2} struggle to accurately identify the edges of the table, as illustrated in \Cref{fig:comparing_table_masks}.  
One major challenge with depth-based models is their difficulty in distinguishing between the top surface and the side of the table, leading to ambiguous table edge detection.  
Meanwhile, SAM tends to over-segment the table, producing multiple disjointed submasks instead of a single cohesive table mask.  
This fragmentation requires additional post-processing to merge the individual segments, adding complexity to the pipeline.  

\begin{figure}[ht]
    \centering
    \includegraphics[width=\linewidth]{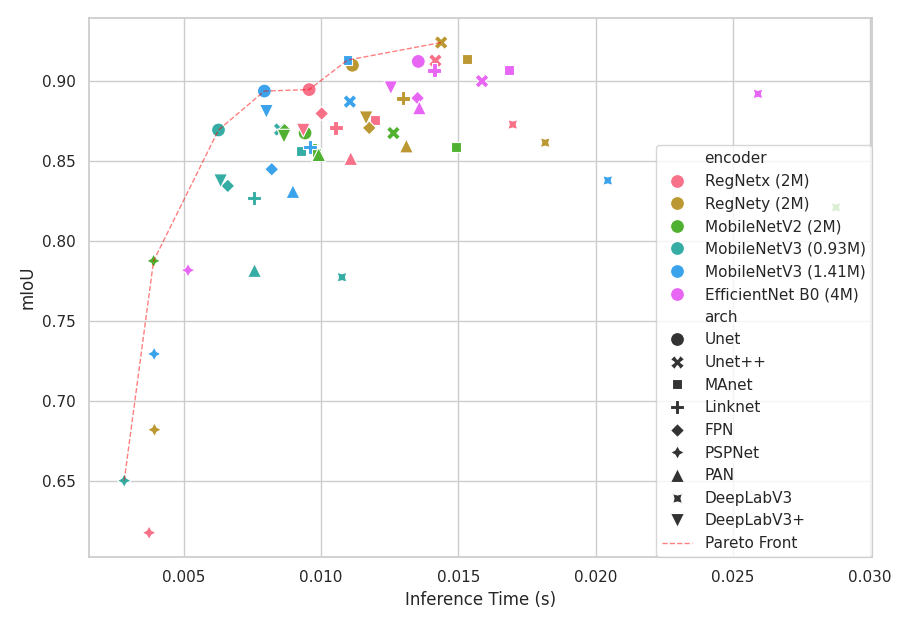}
    \caption{Benchmark of the table segmentation models for different model architecture and encoders. The Pareto from is draw to show the optimal models with regards to mIoU and inference time.}
    \label{fig:mask_benchmark}
\end{figure}

Given these limitations, we opted to fine-tune existing segmentation networks on our custom dataset using the DICE loss.  
The Segmentation Models (PyTorch) library~\cite{Iakubovskii:2019} provided a flexible framework for experimenting with different architectures and loss functions efficiently.  
We evaluated a range of segmentation architectures, including DeepLabV3~\cite{deeplabv3}, DeepLabV3+~\cite{deeplabv3plus}, U-Net~\cite{Unet}, U-Net++~\cite{zhou2018}, PSPNet~\cite{pspnet}, MANet~\cite{manet}, LinkNet~\cite{linknet}, FPN~\cite{fpn}, and PAN~\cite{pan}.  
Additionally, we tested various backbone networks to assess their impact on segmentation accuracy.  
The performance of these models is summarized in \Cref{fig:mask_benchmark}.  

\section{Extension of our method to tennis}
\label{sec:tennis_extension}

While table tennis and tennis differ in scale and equipment, their underlying ball dynamics share significant similarities.
Both sports involve a ball that is required to bounce once on the playing surface, which contrasts with sports like badminton, where the shuttlecock does not bounce. 
The core principles of ball motion and bounce dynamics, therefore, remain relatively consistent between the two sports.

For tennis, existing camera calibration methods are well-established~\cite{farin2004a,han2006}.
These methods typically rely on standard computer vision techniques, such as color filtering and the Hough line transform, to accurately detect court lines. 
The intersections of these lines are then used to calibrate the camera. 
Compared to table tennis, this calibration process is more straightforward due to reduced player occlusion and a larger number of visible features on the court.

Adapting our 3D reconstruction method from table tennis to tennis primarily involves adjusting the physical parameters that govern the ball's motion. 
The tennis environment introduces additional complexity due to the variability in court surfaces—namely, grass, clay, and hard courts (e.g., cement). 
Each surface type influences the ball's behavior through its unique coefficient of restitution ($COR$) and friction coefficient ($\mu$).
For instance, grass courts have a lower $COR$ and friction coefficient, resulting in a lower and faster bounce compared to clay courts, where higher friction increases ball spin and reduces speed post-bounce.
Hard courts typically provide a balance between these extremes. 
We provide these tennis constants in \Cref{tab:tennis_constants}.

\begin{table}
    \centering
    \begin{tabular}{|c|c|}
        \hline
        \textbf{Variable} & \textbf{Tennis} \\
        \hline
         \rowcolor[gray]{0.9}$m$ [kg] & \num{5.7e-2} \\
         $r$ [m] & 0.033 \\
         \rowcolor[gray]{0.9}$\mu$ &  0.55 / 0.9 $^*$ \\
         $k_{COR}$ & 0.68 / \num{0.85}$^*$ \\
         \hline
    \end{tabular}
    \caption{Physical constants used for tennis. $^*$ We distinguish between grass and clay courts, which have different dynamics.}
    \label{tab:tennis_constants}
\end{table}

As for the aerodynamics, the fuzzy surface of the tennis ball makes the dynamics much more complex.
There are multiple formulation for the drag and mangus forces~\cite{robinson2018,wang2019}. 
We chose to use the convention from ~\cite{robinson2018}.
The drag force is then defined as:
\begin{equation}
    \bm{F_D} = \frac{1}{2} C_{D} \rho \pi r^2 \|\bm{v}\| \bm{v}
\label{eq:F_magnus}
\end{equation}
where:
\begin{equation}
\begin{aligned}
C_{\mathrm{D}}= & 0.6204-9.76 \times 10^{-4}(\bm{v}-50) \\
& +\left[1.027 \times 10^{-4}-2.24 \times 10^{-6}(\bm{v}-50)\right] \bm{\omega},
\end{aligned}
\end{equation}
And the Magnus force is defined as:
\begin{equation}
    \bm{F_M} = \frac{1}{2} C_{M} \rho \pi r^2 \|\bm{v}\| \frac{\bm{\omega} \times \bm{v}}{\|\bm{\omega}\|}
\label{eq:F_magnus}
\end{equation}
where:
\begin{equation}
C_{M} = \|\bm{\omega}\| \left( 4.68 \times 10^{-4} - 2.10 \times 10^{-5} (\|\bm{v}\| - 50) \right)
\label{eq:Cm_tennis}
\end{equation}
With the updated \ac{ODE}, the same 3D ball trajectory method can be used.


\begin{figure*}
    \centering
    \begin{tabular}{cccc}
        \subfloat{\includegraphics[width=0.32\textwidth]{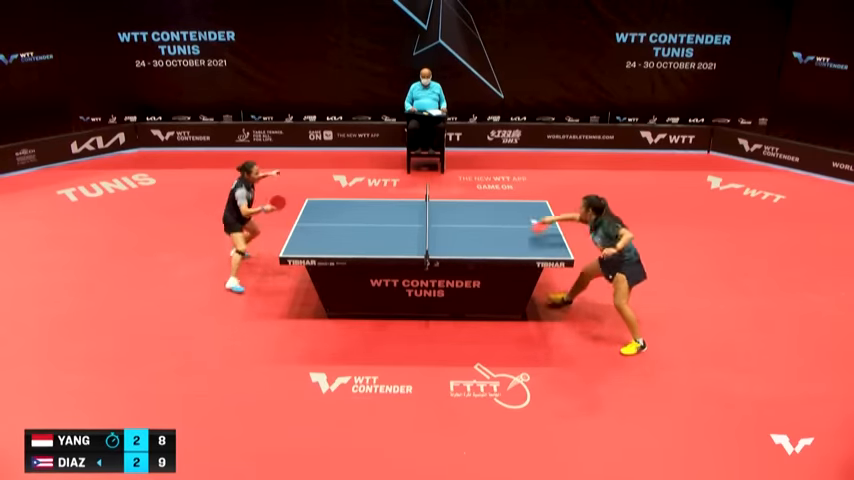}} &
        \subfloat{\includegraphics[width=0.32\textwidth]{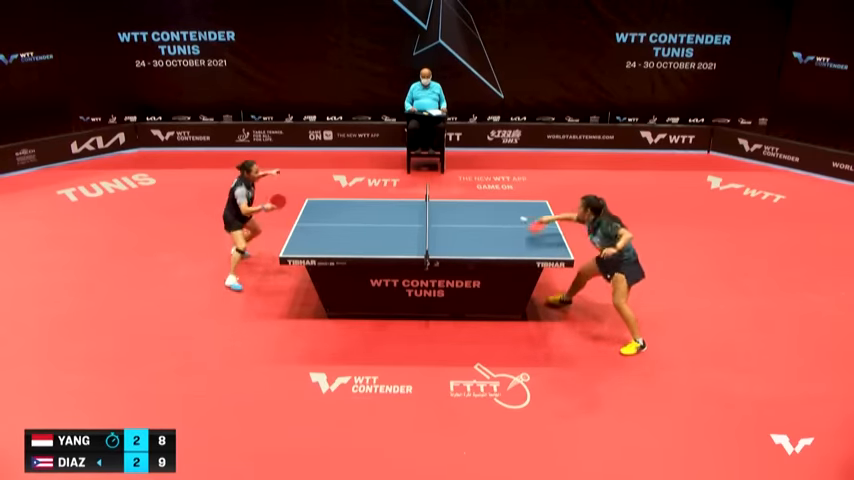}} &
        \subfloat{\includegraphics[width=0.32\textwidth]{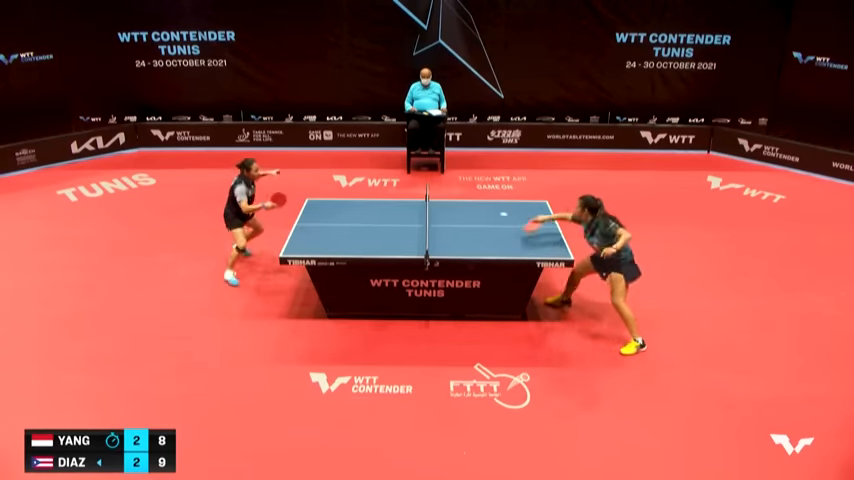}} \\
        
        \subfloat{\includegraphics[width=0.32\textwidth]{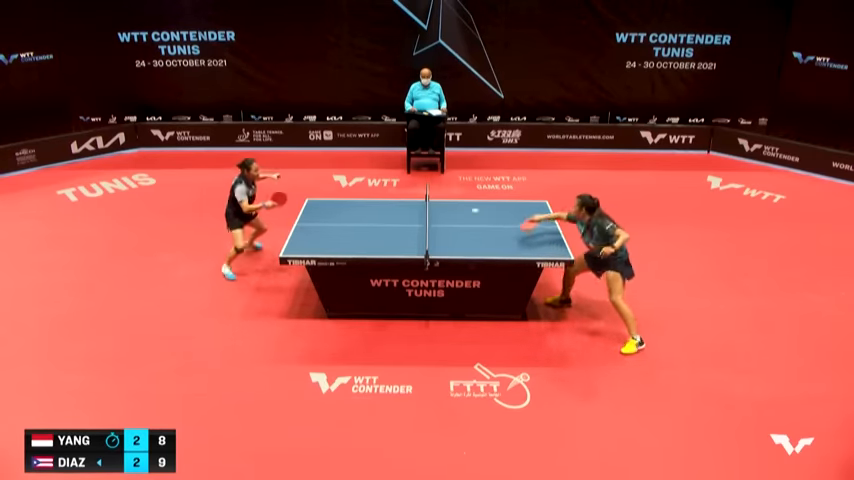}} &
        \subfloat{\includegraphics[width=0.32\textwidth]{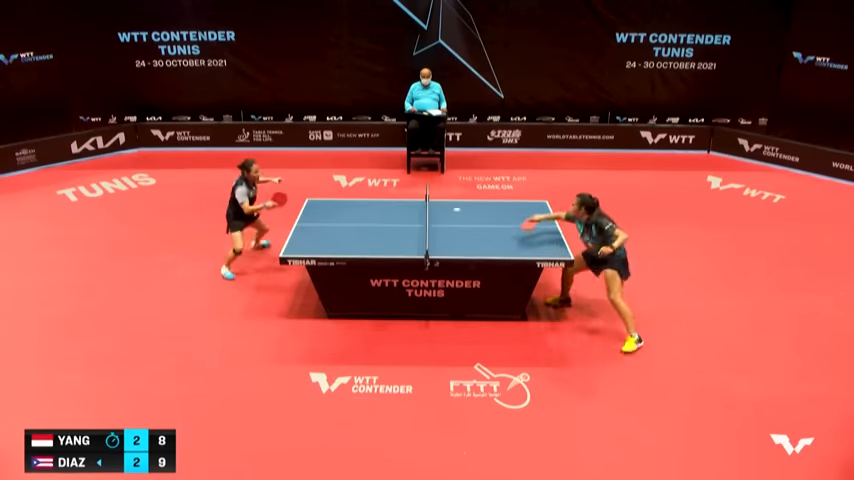}} &
        \subfloat{\includegraphics[width=0.32\textwidth]{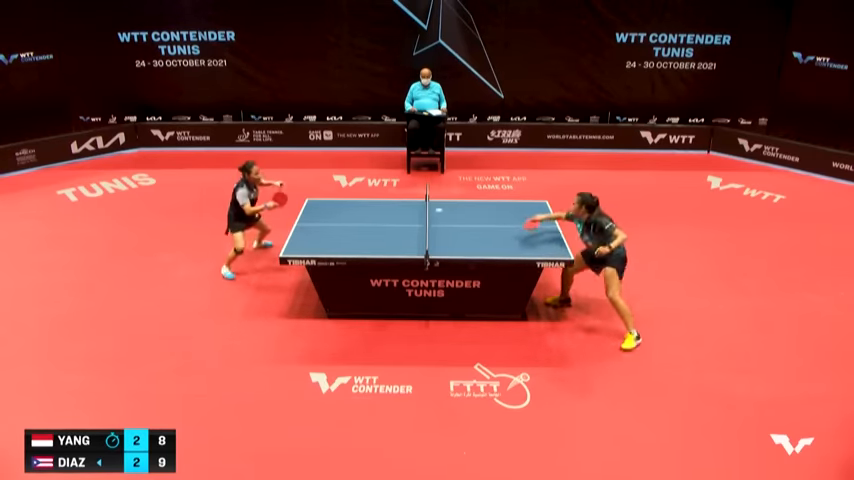}} \\
        
        \subfloat{\includegraphics[width=0.32\textwidth]{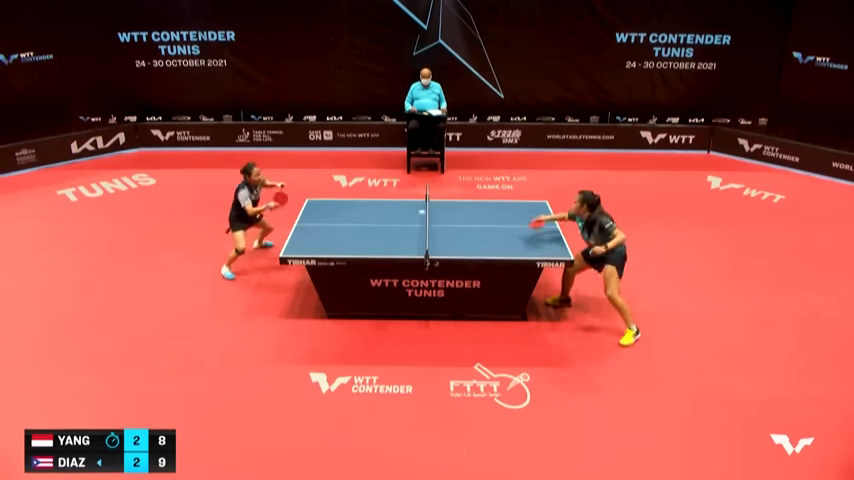}} &
        \subfloat{\includegraphics[width=0.32\textwidth]{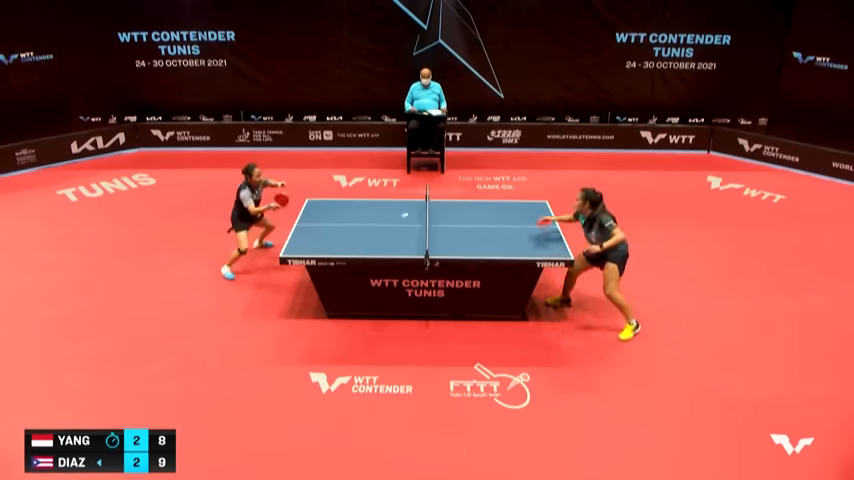}} &
        \subfloat{\includegraphics[width=0.32\textwidth]{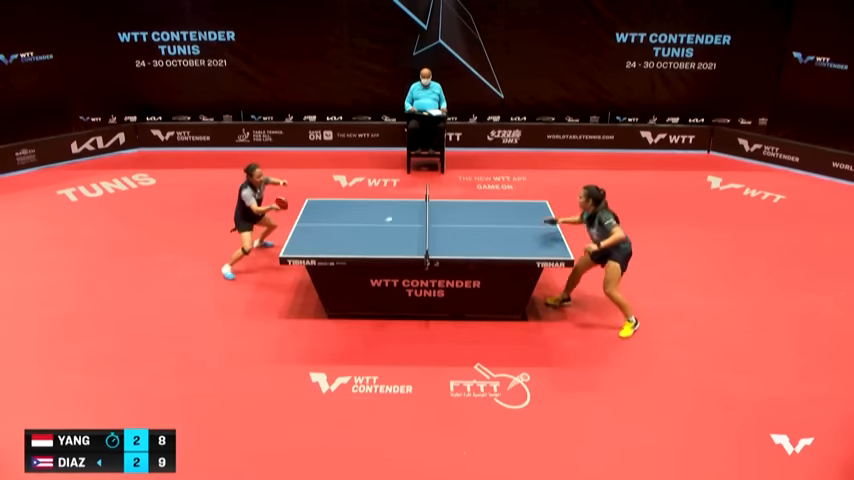}} \\
        
        \subfloat{\includegraphics[width=0.32\textwidth]{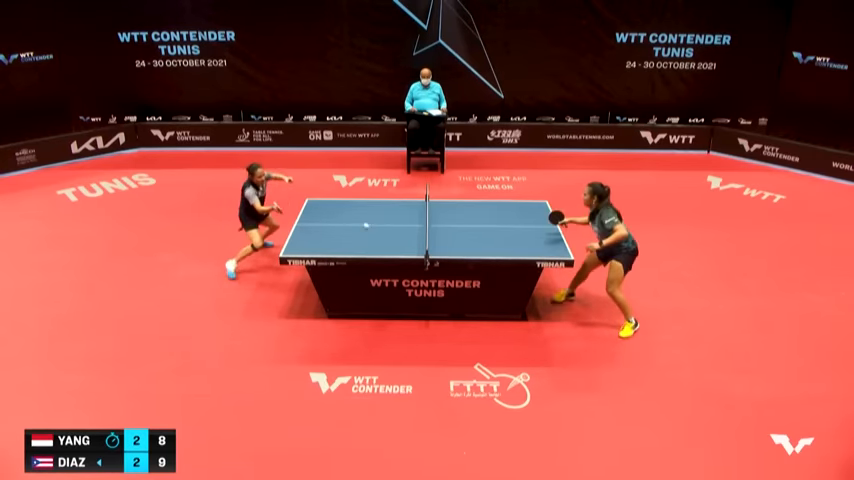}} &
        \subfloat{\includegraphics[width=0.32\textwidth]{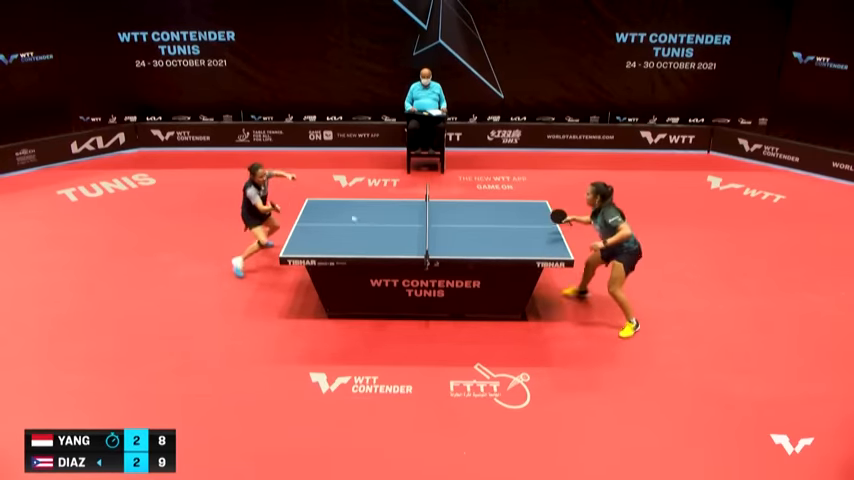}} &
        \subfloat{\includegraphics[width=0.32\textwidth]{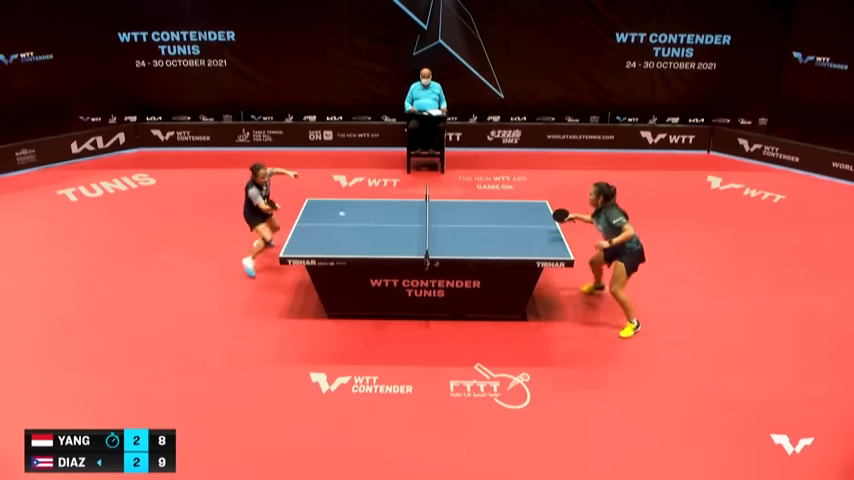}} \\
        
        \subfloat{\includegraphics[width=0.32\textwidth]{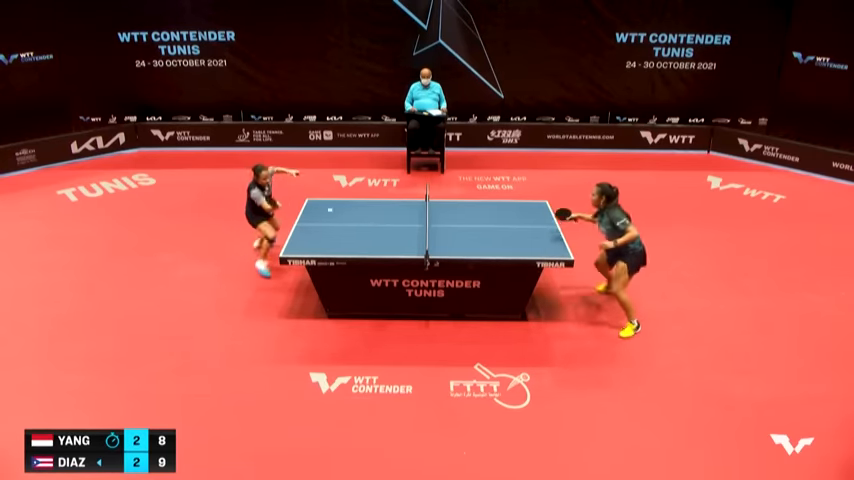}} &
        \subfloat{\includegraphics[width=0.32\textwidth]{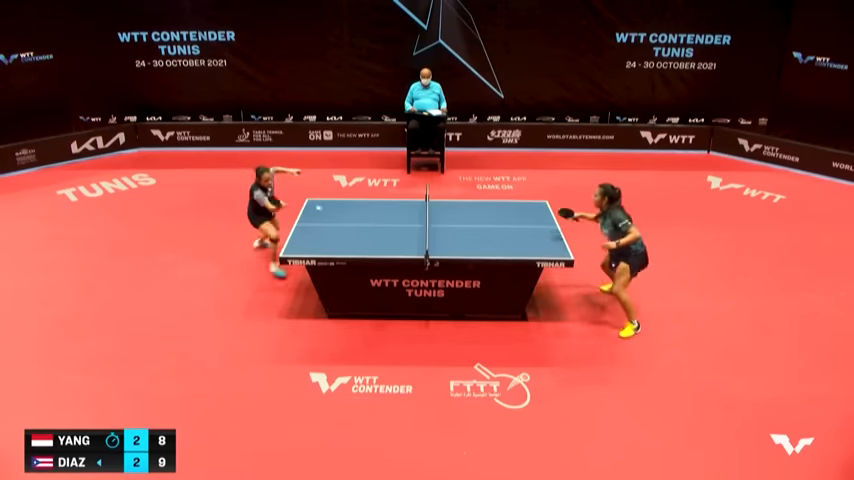}} &
        \subfloat{\includegraphics[width=0.32\textwidth]{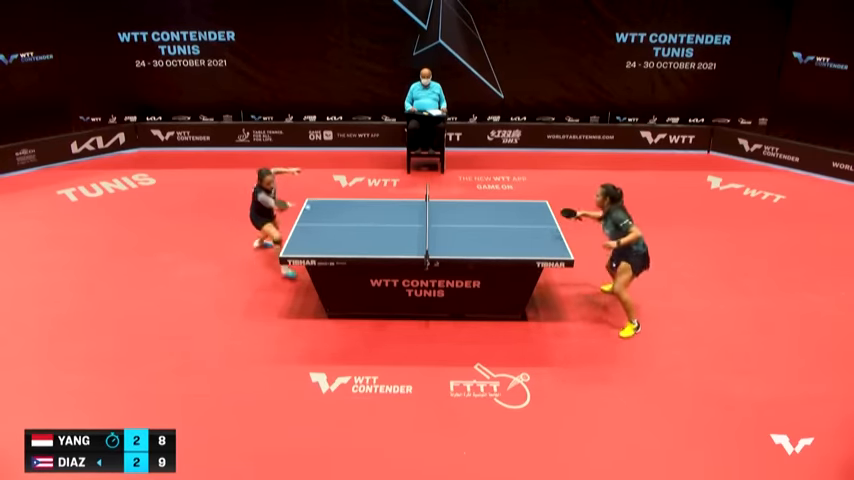}} &
    \end{tabular}

    \begin{subfigure}{0.31\linewidth}
        \includegraphics[width=\linewidth]{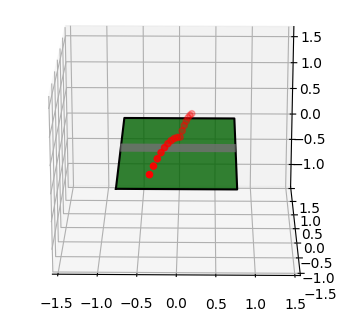}
        \caption*{Backview}
    \end{subfigure}
    \hfill
    \begin{subfigure}{0.31\linewidth}
        \includegraphics[width=\linewidth]{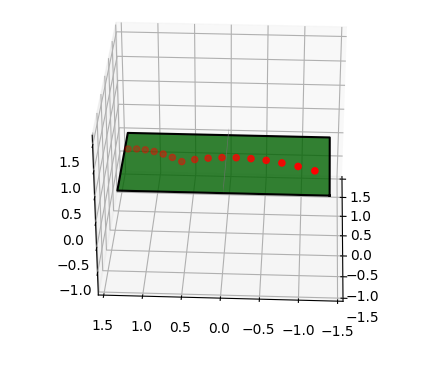}
        \caption*{Sideview}
    \end{subfigure}
    \hfill
        \begin{subfigure}{0.31\linewidth}
        \includegraphics[width=\linewidth]{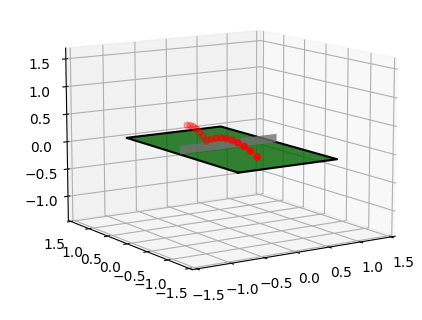}
        \caption*{Oblic view}
    \end{subfigure}
    

    \caption{(Top)Input frames used to reconstruct the 3D ball trajectory. They are sorted in chronological order. (Bottom) 3D reconstruction results obtained from these frames for different points of view.}
    \label{fig:traj_grid}
\end{figure*}

\begin{figure*}
    \centering
    \begin{tabular}{ccc}
        \subfloat{\includegraphics[width=0.31\textwidth]{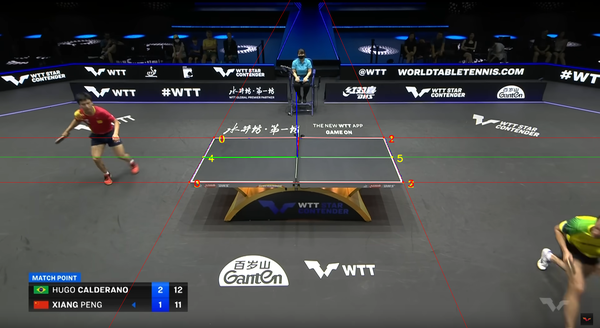}} &
        \subfloat{\includegraphics[width=0.31\textwidth]{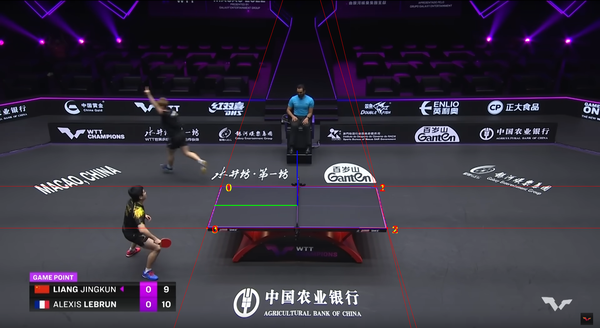}} &
        \subfloat{\includegraphics[width=0.31\textwidth]{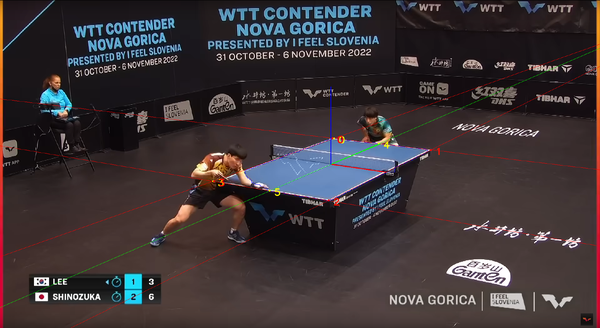}} \\
        
        \subfloat{\includegraphics[width=0.31\textwidth]{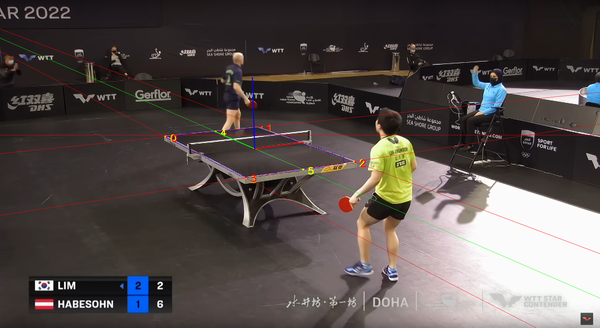}} &
        \subfloat{\includegraphics[width=0.31\textwidth]{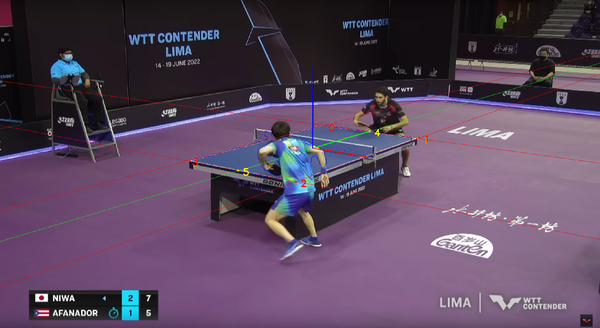}} &
        \subfloat{\includegraphics[width=0.31\textwidth]{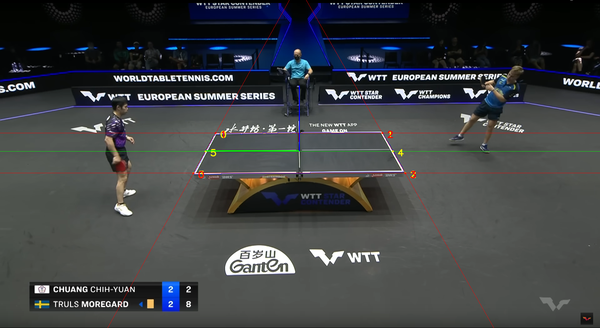}} \\
        
        \subfloat{\includegraphics[width=0.31\textwidth]{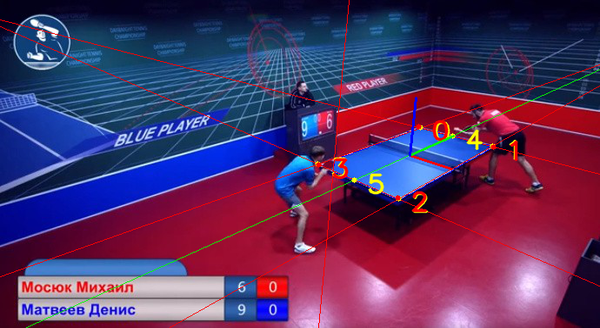}} &
        \subfloat{\includegraphics[width=0.31\textwidth]{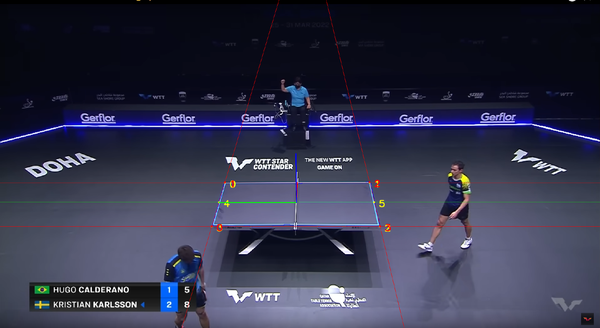}} &
        \subfloat{\includegraphics[width=0.31\textwidth]{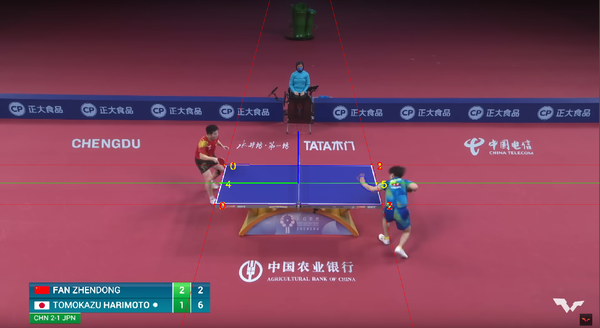}} \\
        
        \subfloat{\includegraphics[width=0.31\textwidth]{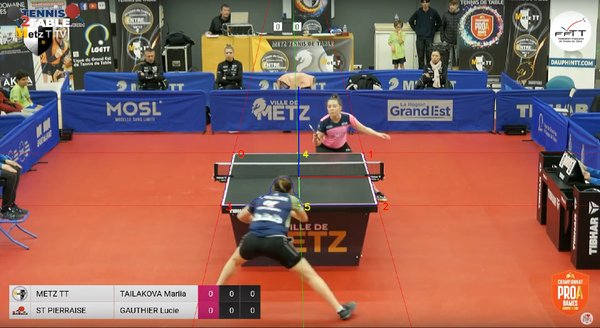}} &
        \subfloat{\includegraphics[width=0.31\textwidth]{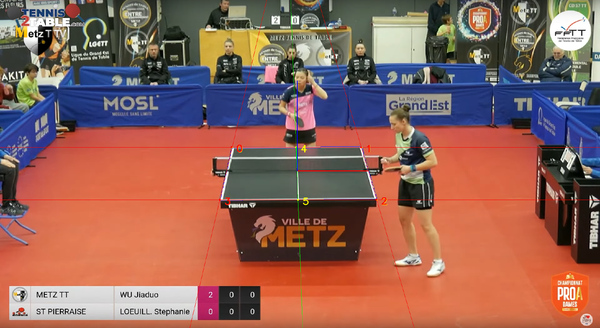}} &
        \subfloat{\includegraphics[width=0.31\textwidth]{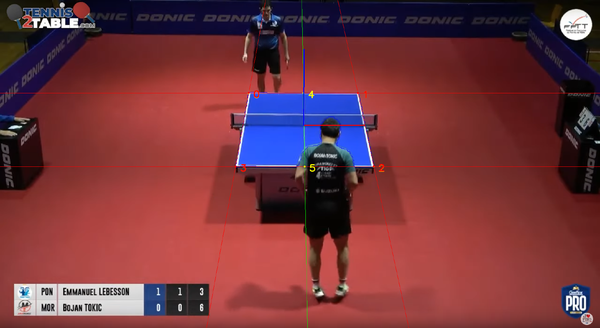}} \\
        
        \subfloat{\includegraphics[width=0.31\textwidth]{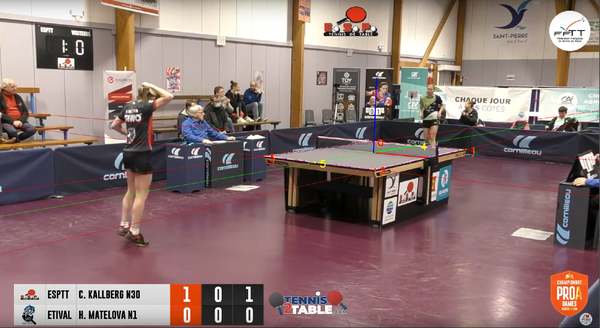}} &
        \subfloat{\includegraphics[width=0.31\textwidth]{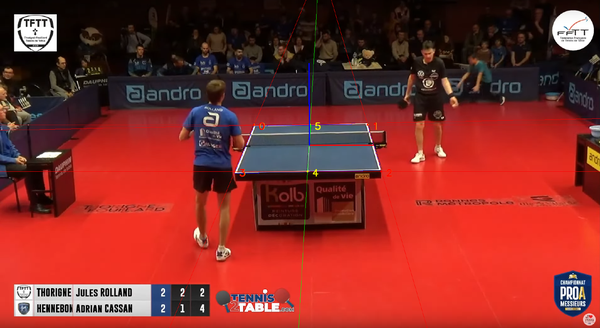}} &
        \subfloat{\includegraphics[width=0.31\textwidth]{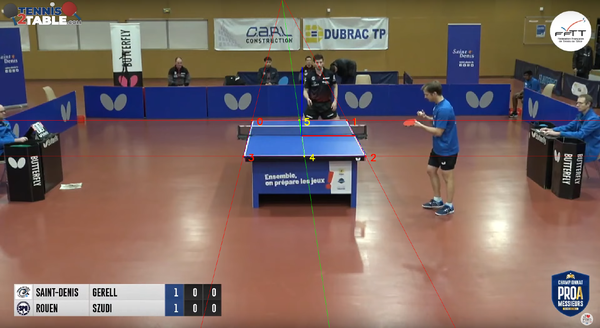}} \\
    \end{tabular}
    \caption{Camera calibration in different contexts. The red lines are the potential detected table edges. The green lines are the potential midlines. The numbers represent the feature number.}
    \label{fig:calib_grid}
\end{figure*}

\begin{figure*}
    \centering
    \begin{tabular}{cc}
        \subfloat{\includegraphics[width=0.45\textwidth]{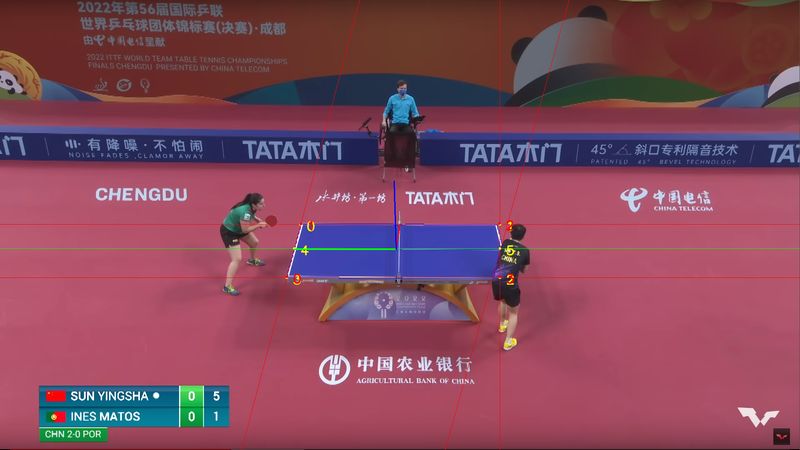}} &
        \subfloat{\includegraphics[width=0.45\textwidth]{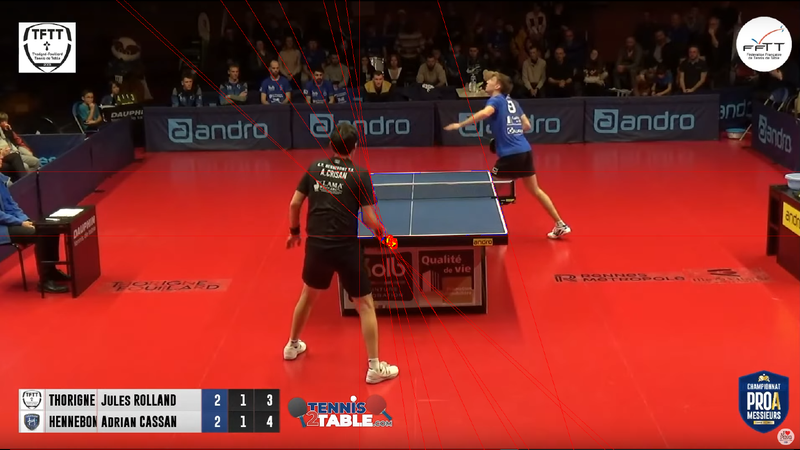}} \\
        \subfloat{\includegraphics[width=0.45\textwidth]{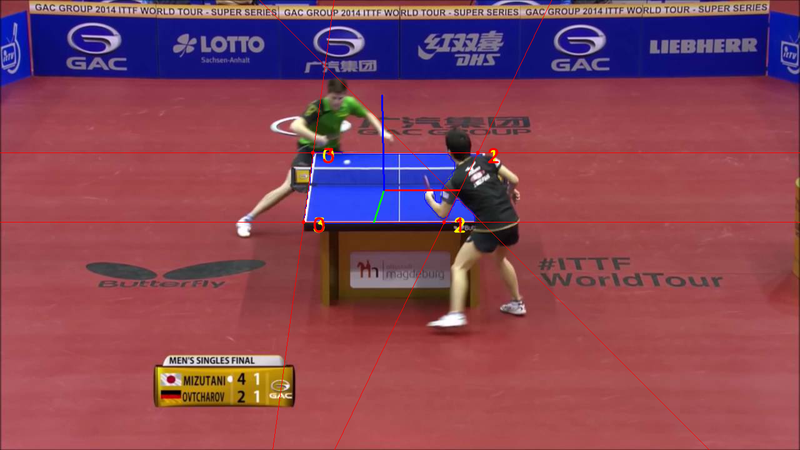}} &
        \subfloat{\includegraphics[width=0.45\textwidth]{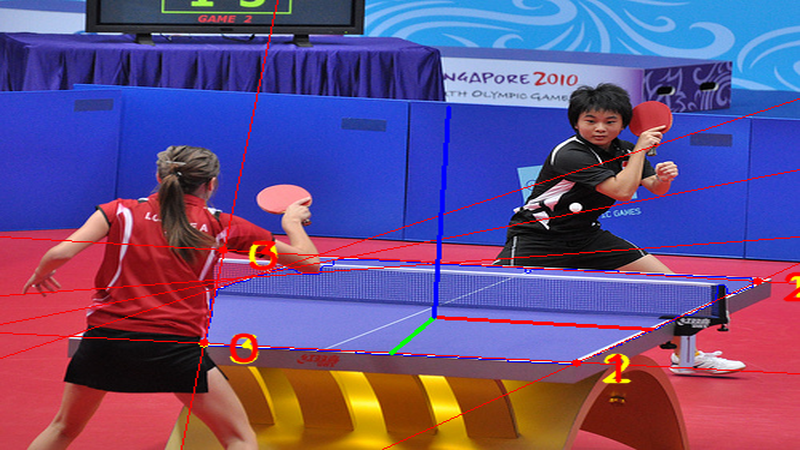}} 
    \end{tabular}
    \caption{Camera calibration failures mostly because of obstruction from the table tennis player. The red lines are the potential detected table edges. The numbers represent the feature number. Most misdetections are due to the player's body being detected as the table edge.}
    \label{fig:calib_grid}
\end{figure*}

\end{document}